\PassOptionsToPackage{unicode}{hyperref}
\PassOptionsToPackage{hyphens}{url}
\documentclass[
]{article}
\usepackage{amsmath,amssymb}
\usepackage{lmodern}
\usepackage{iftex}
\ifPDFTeX
  \usepackage[T1]{fontenc}
  \usepackage[utf8]{inputenc}
  \usepackage{textcomp} 
\else 
  \usepackage{unicode-math}
  \defaultfontfeatures{Scale=MatchLowercase}
  \defaultfontfeatures[\rmfamily]{Ligatures=TeX,Scale=1}
\fi
\IfFileExists{upquote.sty}{\usepackage{upquote}}{}
\IfFileExists{microtype.sty}{
  \usepackage[]{microtype}
  \UseMicrotypeSet[protrusion]{basicmath} 
}{}
\makeatletter
\@ifundefined{KOMAClassName}{
  \IfFileExists{parskip.sty}{%
    \usepackage{parskip}
  }{
    \setlength{\parindent}{0pt}
    \setlength{\parskip}{6pt plus 2pt minus 1pt}}
}{
  \KOMAoptions{parskip=half}}
\makeatother
\usepackage{xcolor}
\usepackage{longtable,booktabs,array}
\usepackage{calc} 
\usepackage{etoolbox}
\makeatletter
\patchcmd\longtable{\par}{\if@noskipsec\mbox{}\fi\par}{}{}
\makeatother
\IfFileExists{footnotehyper.sty}{\usepackage{footnotehyper}}{\usepackage{footnote}}
\makesavenoteenv{longtable}
\usepackage{graphicx}
\makeatletter
\def\maxwidth{\ifdim\Gin@nat@width>\linewidth\linewidth\else\Gin@nat@width\fi}
\def\maxheight{\ifdim\Gin@nat@height>\textheight\textheight\else\Gin@nat@height\fi}
\makeatother
\setkeys{Gin}{width=\maxwidth,height=\maxheight,keepaspectratio}
\makeatletter
\def\fps@figure{htbp}
\makeatother
\ifLuaTeX
  \usepackage{selnolig}  
\fi
\IfFileExists{bookmark.sty}{\usepackage{bookmark}}{\usepackage{hyperref}}
\IfFileExists{xurl.sty}{\usepackage{xurl}}{} 
\hypersetup{
  hidelinks,
  pdfcreator={LaTeX via pandoc}}

\author{}
\date{}

\begin{document}

\begin{center}

\textbf{Gamified AI Approch for Early Detection of Dementia}

\textsuperscript{1}Paramita Kundu Maji\textsuperscript{*},
\textsuperscript{2}Soubhik Acharya, \textsuperscript{3}Priti Paul,
\textsuperscript{4}Sanjay Chakraborty\textsuperscript{*},
\textsuperscript{5}Saikat Basu

\textsuperscript{1234}Department of Computer Science and Engineering,
Techno International New Town, Kolkata, India

\textsuperscript{4}Department of Computer and Information Science,
Linkoping University, Sweden

\textsuperscript{5}Department of Computer Science and Engineering,
Maulana Abul Kalam Azad University, Kolkata, India

Email: paramita.kundu.maji@tint.edu.in\textsuperscript{*},
schakraborty770@gmail.com\textsuperscript{*}

\end{center}

\emph{\textbf{Abstract:} This paper aims to develop a new deep
learning-inspired gaming approach for early detection of dementia. This
research integrates a robust convolutional neural network (CNN)-based
model for early dementia detection using health metrics data as well as
facial image data through a cognitive assessment-based gaming
application. We have collected 1000 data samples of health metrics
dataset from Apollo diagnostic center Kolkata that is labeled as either
``demented'' or ``non-demented'' for the training of MOD-1D-CNN for the
game level 1 and another dataset of facial images containing 1800 facial
data that are labeled as either ``demented'' or ``non-demented'' is
collected by our research team for the training of MOD-2D-CNN model
in-game level 2. In our work, the loss for the proposed MOD-1D-CNN model
is 0.2692 and the highest accuracy is 70.50\% for identifying the
dementia traits using real-life health metrics data. Similarly, the
proposed MOD-2D-CNN model loss is 0.1755 and the highest accuracy is
obtained here 95.72\% for recognizing the dementia status using
real-life face-based image data. Therefore, a rule-based weightage
method is applied to combine both the proposed methods to achieve the
final decision. The MOD-1D-CNN and MOD-2D-CNN models are more
lightweight and computationally efficient alternatives because they have
a significantly lower number of parameters when compared to the other
state-of-the-art models. We have compared their accuracies and
parameters with the other state-of-the-art deep learning models.}


\textbf{1. INTRODUCTION}

A major threat to public health in a period of global population aging
is the development of dementia. Effective intervention and treatment
require early diagnosis of cognitive impairment. Accessibility and
involvement are typically lacking in traditional diagnostic techniques
{[}1-3{]}. This work presents a novel method for identifying early
dementia that draws inspiration from gamified cognitive evaluation and
deep learning concepts {[}4-6{]}. Through the use of a two-tiered memory
card matching game, users---demented or not---participate in a dynamic
evaluation. If users complete the first level within the allotted click
threshold and time frame, they can go to the second level; if not, they
are prompted to provide pertinent health parameters and facial image
data. Once the first level has been completed successfully, users are
successfully redirected to the second level. If both levels are
completed successfully, a message congratulating them on moving through
the cognitive assessment levels appears. If users cannot complete the
first level of prediction within the designated click threshold and time
limit, they are prompted to provide vital health indicators, including
age, weight, heart rate, blood oxygen levels, body temperature, weight,
and diabetes status, to aid with the first level of prediction. After
providing these details, gamers are granted an additional opportunity to
go to the second level. Users who do not meet the click threshold on the
second level are prompted to capture an image of their face during
gameplay. A deep-learning model then examines this image to look for
signs of dementia. The user is presented with definitive results at the
end of this interactive process, which has reshaped the field of early
dementia diagnosis by combining artificial intelligence, health
parameter input, and game-based involvement.

By developing a gamified cognitive evaluation tool based on deep
learning principles, this research aims to tackle the urgent problem of
early dementia identification. The main issue is the low interest and
accessibility of conventional diagnostic techniques, which calls for a
new strategy that seamlessly combines health parameter input, face
picture analysis, and a two-tiered memory card matching game. The main
research question that motivates this investigation is whether this
novel approach---which combines artificial intelligence and game-based
interaction---can offer a practical and efficient way to detect
cognitive decline in its early stages, thereby enhancing intervention
tactics and the care given to those who are at risk of dementia. This
work aims to determine the effectiveness of a gamified cognitive
assessment system in the early detection of dementia by using a
two-tiered memory card-matching game. Additionally, the accuracy and
reliability of the deep learning model in predicting dementia based on
user-provided health parameters and facial images will be assessed. The
work's central premise is that combining a dynamic gaming framework with
deep learning-based facial image analysis and health parameter input
would result in a reliable and approachable way to spot cognitive
impairment early on. By achieving these goals, this research hopes to
provide insightful information on how cutting-edge technology might
transform current approaches to early dementia detection.

There is some interesting literature available in this domain. The
literature on dementia detection highlights the critical need for early
identification, given the limitations of traditional diagnostic
approaches regarding accessibility and engagement. Notable studies
explore various innovative methodologies, such as Kherchouche et al.
{[}7{]} Deep Learning framework for early Alzheimer's disease detection
using Proton Magnetic Resonance Spectroscopy (1H-MRS) data, emphasizing
concerns about data limitations and interpretability. Alex et al.
{[}8{]} propose a cost-effective and non-invasive approach for early AD
detection using speech data, acknowledging challenges in dataset
diversity and model interpretability. Qayyum et al. {[}9{]} employ EEG
data analysis to discern resting and cognitive states, achieving notable
achievements but recognizing limitations in sample size and model
interpretability. Gamification emerges as a promising avenue, with
Paliokas et al. {[}10{]} introducing a gamified social platform for
dementia care, acknowledging challenges in accessibility and
generalizability. Zhang and Chignell {[}11{]} explore cognitive
assessment games for people with dementia, addressing concerns about
generalizability and ethical considerations. Chi et al. {[}12{]}
introduce the ``Smart Thinker'' game suite, showing cognitive
improvement but recognizing limitations in study duration and sample
size. Mora et al. {[}13{]} The PNH tool targets cognitive training,
addressing challenges in generalizability and ethical user privacy
practices. Technological interventions include Paletta et al. {[}14{]}
PLAYTIME app, combining eye movement analysis and cognitive exercises,
with concerns about varying efficacy and the need for clinical
validation. Aram et al. {[}15{]} introduce a 2-layer model for early
detection of dementia problems, emphasizing the need for ethical
considerations and diverse dataset validation. Shankle et al. {[}16{]}
differentiate between normal aging and early dementia using machine
learning, highlighting the importance of addressing dataset specificity.
Bidani et al. {[}17{]} present a Deep Learning approach for dementia
detection using MRI scans, facing challenges in dataset homogeneity and
model interpretability. Murugan et al. {[}18{]} DEMNET framework
achieves remarkable accuracy but lacks extensive discussion on
interpretability and broader dataset validation. Table 1 presents
concise and intriguing research gaps drawn from existing literature that
have served as inspiration for our current work.

\begin{center}
\textbf{Table 1:} Research gaps in some key literature works
\end{center}

\begin{longtable}[]{@{}
  >{\raggedright\arraybackslash}p{(\columnwidth - 2\tabcolsep) * \real{0.2985}}
  >{\raggedright\arraybackslash}p{(\columnwidth - 2\tabcolsep) * \real{0.7015}}@{}}
\toprule()
\begin{minipage}[b]{\linewidth}\raggedright

\textbf{Key Literature Works}
\end{minipage} & \begin{minipage}[b]{\linewidth}\raggedright

\textbf{Research Gaps}
\end{minipage} \\
\midrule()
\endhead
{[}7{]} & The major research need is for more potent early detection
techniques for Alzheimer's disease (AD) that use data from proton
magnetic resonance spectroscopy (1H-MRS) to identify biological
alterations in the brain before the onset of structural atrophy. 
\\
{[}8{]} & The research gap is the need for more investigation and
development of robust and explainable multimodal fusion approaches for
speech-based automated detection of Alzheimer's disease (AD), taking
into account issues with reliability, data size, and model
explainability. \\
{[}9{]} & The area of unmet research need is the investigation of deep
learning methods, specifically 1D convolutional neural networks (CNNs),
for precise classification and evaluation of cognitive load from EEG
data. Specifically, the study should concentrate on the consistent
pattern of alpha brain waves in differentiating between resting and
cognitive states. \\
{[}10{]} & The area of unmet research need is the need for more
investigation and the creation of gamification systems specifically
designed for dementia patients and their carers in social networking
settings. \\
{[}11{]} & The area of unmet research need is the creation of a thorough
framework for the efficient use of cognitive assessment games made
especially for dementia patients. \\
{[}12{]} & The research gap lies in the area of computerized serious
games, such as ``Smart Thinker'', that target memory and attention
abilities in older persons to improve cognitive functioning. This is
especially important when it comes to managing and preventing
dementia. \\
{[}13{]} & The research gap concerns the need for novel and compelling
techniques to improve long-term adherence and the efficacy of
personalized cognitive impairment prevention treatments in older
persons. Examples of such initiatives include gamified
crowdsourcing-inspired programs like Preventive Neuro Health (PNH). \\
{[}14{]} & The research gap is the lack of longitudinal data on changing
mental processes in dementia, which makes it difficult to provide
personalized therapy and prolonged freedom at home. \\
{[}15{]} & The research gap relates to the requirement for machine
learning-based early diagnosis models for dementia that are both
effective and efficient. These models can alleviate the load on patients
and healthcare systems by streamlining diagnostic procedures. \\
{[}16{]} & The research gap relates to the need for machine learning
techniques that can use extensive datasets to discern between early
symptoms of dementia and normal brain aging, perhaps offering more
accessible and accurate diagnostic tools than previous methods. \\
{[}17{]} & The research gap is to meet the demand for better diagnostic
instruments in the evaluation of neurodegenerative diseases, a novel
method utilizing Deep Convolutional Neural Network (DCNN) and Transfer
Learning models has been developed to reliably identify and categorize
dementia using MRI brain images. \\
{[}18{]} & The research gap is the need for better Convolutional Neural
Network (CNN) models-based diagnostic processes for Alzheimer's Disease
(AD), which address class imbalance and increase sensitivity and
precision in MRI image recognition. \\
\bottomrule()
\end{longtable}

This work adds substantially to the corpus of information by addressing
the urgent need for novel and approachable techniques in the early
identification of dementia. To create a practical and efficient method
for detecting cognitive decline early on, we consider implementing a
gamified cognitive evaluation system. This system can help assess
cognitive function and identify potential signs of decline. By gamifying
the evaluation process, we can engage users and make the assessment more
enjoyable, increasing their willingness to participate.The scope
comprises developing a two-level memory game that combines gaming, deep
learning models, face image analysis, and health parameter input for an
all-encompassing cognitive evaluation. The work does, however, have many
drawbacks, such as a limited sample size, potential biases related to
user participation in gamified environments, and the requirement for
additional validation in a variety of demographics. Individual variances
in user behavior can impact the results and might not fully represent
the range of characteristics associated with dementia. The methodology
includes developing and implementing a gamified cognitive evaluation
system, using deep learning algorithms to detect dementia, and using a
quantitative approach to examine user performance data. Important
elements of the procedure include data validation, model accuracy
evaluation, and analysis of user input to improve the system's
dependability and user-friendliness. In the end, our research aims to
transform methods for detecting dementia, encouraging early
intervention, and enhancing patient outcomes in the context of proactive
dementia care.

The rest of this paper is organized as follows. Section 2 gives a
detailed description of the proposed methodology. In section 2, the
different levels of the proposed game, MOD-1D-CNN and MOD-2D-CNN
algorithms \& their architectures are explained. An experimental setup,
feature extraction, and mapping strategies are provided in section 3.
The result analysis and discussion along with a brief comparison is
given in Section 4. Section 5 discusses the conclusions of the work.

\textbf{2. PROPOSED METHODOLOGY}

\textbf{2.1 Levels of Proposed Game}

Cognitive talents comprise a wide range of human abilities, including
psychomotor, olfactory, tactile, visual processing, reasoning, reading,
writing, and others {[}19{]}. Executive Functions (EFs) are cognitive
processes that exert influence over other abilities and regulate their
functioning {[}20{]}. Disruptions in cognitive processes can result in
the development of social and mental health issues {[}21{]}. Cognitive
assessment is a valuable tool for evaluating cognitive impairment, which
refers to a lack of knowledge, cognitive functioning, or sound judgment.
To perform cognitive assessments, a variety of popular techniques are
used.Each assessment is meticulously designed to assess specific
neuropsychological domains, including memory, language, executive
function, abstract reasoning, attention, and visuospatial skills
{[}22{]}. This innovative cognitive assessment game measures critical
cognitive abilities such as memory, decision-making, and flexibility,
which are essential for the early identification of dementia. It has two
levels to evaluate various cognitive abilities and is created with a
hypertext markup language (HTML), cascading style sheets (CSS), and
JavaScript. The game provides a sophisticated diagnostic experience,
ranging from the ``Basic Match'' level, which evaluates memory
retention, to the ``Ultimate Test,'' which evaluates adaptability and
strategic thinking {[}23-24{]}. This ground-breaking tool goes beyond
traditional gaming by providing a sophisticated diagnostic method and
possibly identifying early cognitive deviations through
technology-driven assessments. In line with the overall goals of the
research, the proposed Level 1 of the ``Basic Match'' game is intended
to serve as a fundamental cognitive evaluation instrument. This
grid-layout game with face-down cards that hold concealed images is
created with hypertext markup language (HTML), Cascading Style Sheets
(CSS), and JavaScript. JavaScript logic is in charge of controlling key
operations like card flipping, pair matching detection, and game
progression. Using fundamental JavaScript methods for element selection,
sound setup, card order randomization, and memory retention countdown,
the Level 1 structure follows card-matching guidelines. Card faces are
momentarily revealed and hidden as users begin the game by using the
`clicking()' method. Select cards can be flipped concurrently by users
using the `enableClick()' method, and a pair detection algorithm is used
by the `match()' function. Potential user challenges are addressed with
an inventive feature. The game will automatically forward players to the
next page for cognitive prediction or assessment if their clicks have
crossed the limits of the number of click counts, which is the click
threshold. Using window.location.href, a JavaScript listener initiates a
redirection function and keeps track of the number of clicks. An
on-screen notification informing that you have beyond the click
threshold ends the game along with the redirection. Before the user
reaches the prediction page, a confirmation popup makes sure they have
acknowledged it. The prediction page functions as a thorough platform,
gathering essential input data for 1D Convolution Neural Network
(CNN)-based dementia prediction. Users provide details about their body
temperature, heart rate, weight, age, diabetes, and blood oxygen level.
Using the algorithms of 1D Convolution Neural Network (CNN) to produce
predicted insights about cognitive health and possible early indicators
of dementia, this data powers an assessment methodology. After the
inputs are validated by the model they are again redirected to play the
next level of the game which is the Ultimate test for the cognitive
assessment. In Level 1, the click threshold is set at 36, the countdown
timer is 2 minutes, and the show timer for remembering card positions is
5 seconds. The forecasts are in line with fundamental ideas that
highlight memory, judgment, and flexibility as critical components of
early dementia identification. The suggested cognitive evaluation
approach is made more useful overall by this smooth transition from
gameplay to cognitive assessment in the Level 1 setting. The proposed
game level 1 is shown in Fig. 1.

\begin{center}
\includegraphics[width=2.57055in,height=2.04484in]{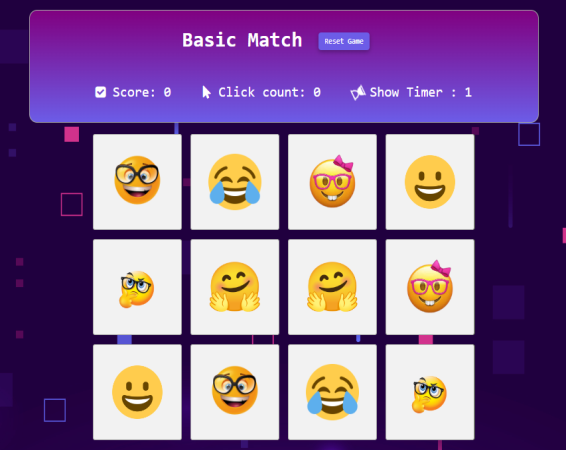}

\textbf{Fig. 1} Basic Match Level 1 of the Proposed Gaming Web
Application
\end{center}

The cognitive evaluation game's Level 2 of the ``Ultimate Test'' is
designed to provide players with a more challenging experience,
introducing a variety of cognitive activities, including grid control,
rotations, swaps, and bomb interactions {[}23-24{]}. The HTML structure
maintains familiarity while enhancing user engagement through the
addition of interactive buttons and a header area. JavaScript scripts
ensure dynamic shuffling of the grid at the beginning of each game,
creating a randomized card layout. To add a layer of complexity, a
countdown timer is implemented, offering a brief memorization period for
players. Clicking on dynamically rotating grid pieces allows players to
match pairs of cards, with the requirement to complete all matches
within a specified click count for a predetermined score. Players who
complete the level within the given click threshold and countdown timer
receive acknowledgment through a message, indicating that they have
successfully passed the cognitive assessment and show no signs of
dementia. However, for those unable to finish within the allotted click
count, the game seamlessly transitions to a facial image capture prompt.
Players are instructed to take a picture of their faces, initiating a 2D
convolutional neural network (CNN) analysis. The click threshold counter
and countdown timer emphasize precision and time sensitivity, ensuring a
challenging cognitive experience. The facial image analysis aims to
predict whether the player exhibits dementia-like symptoms, providing
valuable insights into cognitive health based on facial traits. This
comprehensive approach combines gameplay elements, data input, and
facial image analysis to offer users a thorough cognitive assessment
system. The proposed game level 2 is shown in Fig. 2.

\begin{center}

\includegraphics[width=2.59802in,height=2.14154in]{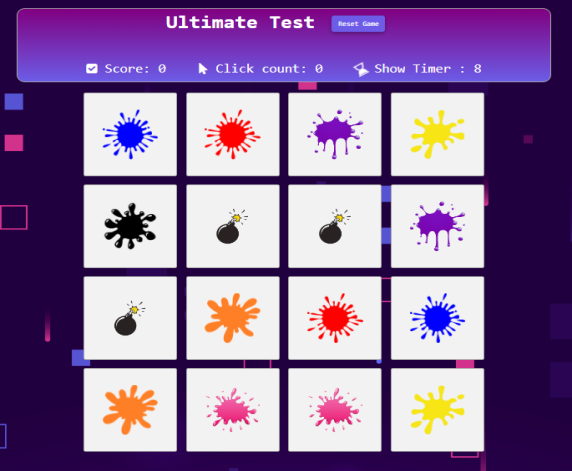}

\textbf{Fig. 2} Ultimate Test Level 2 of the Proposed Gaming Web
Application
\end{center}

\textbf{2.2 Proposed Gaming Strategy}

The procedural overview of the memory game with an integrated dementia
predictor is shown below, which incorporates interactive features,
predictive modeling, and facial image analysis to create a comprehensive
cognitive evaluation system. The proposed algorithm for the level 1 game
is shown below:

\begin{longtable}[]{@{}
  >{\raggedright\arraybackslash}p{(\columnwidth - 8\tabcolsep) * \real{0.0704}}
  >{\raggedright\arraybackslash}p{(\columnwidth - 8\tabcolsep) * \real{0.0640}}
  >{\raggedright\arraybackslash}p{(\columnwidth - 8\tabcolsep) * \real{0.0702}}
  >{\raggedright\arraybackslash}p{(\columnwidth - 8\tabcolsep) * \real{0.0718}}
  >{\raggedright\arraybackslash}p{(\columnwidth - 8\tabcolsep) * \real{0.7235}}@{}}
\toprule()
\multicolumn{5}{@{}>{\raggedright\arraybackslash}p{(\columnwidth - 8\tabcolsep) * \real{1.0000} + 8\tabcolsep}@{}}{%
\begin{minipage}[b]{\linewidth}\raggedright
\textbf{ALGORITHM-1: LEVEL -- 1: BASIC MATCH GAME}
\end{minipage}} \\
\midrule()
\endhead
&
\multicolumn{4}{>{\raggedright\arraybackslash}p{(\columnwidth - 8\tabcolsep) * \real{0.9296} + 6\tabcolsep}@{}}{%
\emph{\textbf{Input:} Game Level}} \\
&
\multicolumn{4}{>{\raggedright\arraybackslash}p{(\columnwidth - 8\tabcolsep) * \real{0.9296} + 6\tabcolsep}@{}}{%
\emph{\textbf{Output:} -User interactions with face-down cards.}

\emph{- Completion of the game.}

\emph{-User input for cognitive prediction (body temperature, heart
rate, weight, age, diabetes, blood oxygen level).}

\emph{-Prediction result using Mod 1D CNN.}

\emph{-Validation feedback for user inputs.}} \\
&
\multicolumn{4}{>{\raggedright\arraybackslash}p{(\columnwidth - 8\tabcolsep) * \real{0.9296} + 6\tabcolsep}@{}}{%
\emph{\textbf{Procedure:}}

\emph{\textbf{Begin:}}} \\
& &
\multicolumn{3}{>{\raggedright\arraybackslash}p{(\columnwidth - 8\tabcolsep) * \real{0.8656} + 4\tabcolsep}@{}}{%
\emph{\textbf{Initialize game components:} setUpGame()}} \\
& &
\multicolumn{3}{>{\raggedright\arraybackslash}p{(\columnwidth - 8\tabcolsep) * \real{0.8656} + 4\tabcolsep}@{}}{%
\emph{\textbf{Start game:} startGame()}} \\
& &
\multicolumn{3}{>{\raggedright\arraybackslash}p{(\columnwidth - 8\tabcolsep) * \real{0.8656} + 4\tabcolsep}@{}}{%
\emph{\textbf{Set level-specific parameters:}}} \\
& & &
\multicolumn{2}{>{\raggedright\arraybackslash}p{(\columnwidth - 8\tabcolsep) * \real{0.7953} + 2\tabcolsep}@{}}{%
\emph{clickThreshold = 36}} \\
& & &
\multicolumn{2}{>{\raggedright\arraybackslash}p{(\columnwidth - 8\tabcolsep) * \real{0.7953} + 2\tabcolsep}@{}}{%
\emph{timerDuration = 2 minutes}} \\
& & &
\multicolumn{2}{>{\raggedright\arraybackslash}p{(\columnwidth - 8\tabcolsep) * \real{0.7953} + 2\tabcolsep}@{}}{%
\emph{showTimerDuration = 5 seconds}} \\
& &
\multicolumn{3}{>{\raggedright\arraybackslash}p{(\columnwidth - 8\tabcolsep) * \real{0.8656} + 4\tabcolsep}@{}}{%
\emph{\textbf{Show timer for memorization:}
showTimer(showTimerDuration)}} \\
& &
\multicolumn{3}{>{\raggedright\arraybackslash}p{(\columnwidth - 8\tabcolsep) * \real{0.8656} + 4\tabcolsep}@{}}{%
\emph{\textbf{while(gameInProgress() and withinTimeLimit(timerDuration))
do}}

\emph{// User interaction loop with timer}} \\
& & &
\multicolumn{2}{>{\raggedright\arraybackslash}p{(\columnwidth - 8\tabcolsep) * \real{0.7953} + 2\tabcolsep}@{}}{%
\emph{\textbf{Allow user to flip cards:} handleCardFlipping()}} \\
& & &
\multicolumn{2}{>{\raggedright\arraybackslash}p{(\columnwidth - 8\tabcolsep) * \real{0.7953} + 2\tabcolsep}@{}}{%
\emph{\textbf{if foundMatch() then} // Check for completed matches}} \\
& & & & \emph{handleMatchedPair()} \\
& & &
\multicolumn{2}{>{\raggedright\arraybackslash}p{(\columnwidth - 8\tabcolsep) * \real{0.7953} + 2\tabcolsep}@{}}{%
\emph{\textbf{end if}}} \\
& & &
\multicolumn{2}{>{\raggedright\arraybackslash}p{(\columnwidth - 8\tabcolsep) * \real{0.7953} + 2\tabcolsep}@{}}{%
\emph{\textbf{if exceedClickThreshold(clickThreshold) then} // Check for
click threshold}} \\
& & & & \emph{initiateCognitivePrediction()} \\
& & & & \emph{displayConfirmationPopup()} \\
& & & & \emph{exitGameLoop()} \\
& & &
\multicolumn{2}{>{\raggedright\arraybackslash}p{(\columnwidth - 8\tabcolsep) * \real{0.7953} + 2\tabcolsep}@{}}{%
\emph{\textbf{end if}}} \\
& &
\multicolumn{3}{>{\raggedright\arraybackslash}p{(\columnwidth - 8\tabcolsep) * \real{0.8656} + 4\tabcolsep}@{}}{%
\emph{\textbf{end while}}} \\
& &
\multicolumn{3}{>{\raggedright\arraybackslash}p{(\columnwidth - 8\tabcolsep) * \real{0.8656} + 4\tabcolsep}@{}}{%
\emph{\textbf{Collect user input for prediction:}
collectUserInput()}} \\
& &
\multicolumn{3}{>{\raggedright\arraybackslash}p{(\columnwidth - 8\tabcolsep) * \real{0.8656} + 4\tabcolsep}@{}}{%
\emph{\textbf{Use Mod 1D CNN for cognitive prediction:}
use1DCNNForCognitivePrediction()}} \\
& &
\multicolumn{3}{>{\raggedright\arraybackslash}p{(\columnwidth - 8\tabcolsep) * \real{0.8656} + 4\tabcolsep}@{}}{%
\emph{\textbf{if inputsValidated() then} // Validate prediction
inputs}} \\
& & &
\multicolumn{2}{>{\raggedright\arraybackslash}p{(\columnwidth - 8\tabcolsep) * \real{0.7953} + 2\tabcolsep}@{}}{%
\emph{redirectToLevel2()}} \\
& &
\multicolumn{3}{>{\raggedright\arraybackslash}p{(\columnwidth - 8\tabcolsep) * \real{0.8656} + 4\tabcolsep}@{}}{%
\emph{\textbf{else}}} \\
& & &
\multicolumn{2}{>{\raggedright\arraybackslash}p{(\columnwidth - 8\tabcolsep) * \real{0.7953} + 2\tabcolsep}@{}}{%
\emph{displayFeedback()}} \\
& &
\multicolumn{3}{>{\raggedright\arraybackslash}p{(\columnwidth - 8\tabcolsep) * \real{0.8656} + 4\tabcolsep}@{}}{%
\emph{\textbf{end if}}} \\
&
\multicolumn{4}{>{\raggedright\arraybackslash}p{(\columnwidth - 8\tabcolsep) * \real{0.9296} + 6\tabcolsep}@{}}{%
\emph{\textbf{end}}} \\
\bottomrule()
\end{longtable}

The proposed algorithm for the level 2 game is shown below:

\begin{longtable}[]{@{}
  >{\raggedright\arraybackslash}p{(\columnwidth - 8\tabcolsep) * \real{0.0699}}
  >{\raggedright\arraybackslash}p{(\columnwidth - 8\tabcolsep) * \real{0.0649}}
  >{\raggedright\arraybackslash}p{(\columnwidth - 8\tabcolsep) * \real{0.0699}}
  >{\raggedright\arraybackslash}p{(\columnwidth - 8\tabcolsep) * \real{0.0516}}
  >{\raggedright\arraybackslash}p{(\columnwidth - 8\tabcolsep) * \real{0.7438}}@{}}
\toprule()
\multicolumn{5}{@{}>{\raggedright\arraybackslash}p{(\columnwidth - 8\tabcolsep) * \real{1.0000} + 8\tabcolsep}@{}}{%
\begin{minipage}[b]{\linewidth}\raggedright
\textbf{ALGORITHM-2: LEVEL -- 2: ULTIMATE TEST GAME}
\end{minipage}} \\
\midrule()
\endhead
&
\multicolumn{4}{>{\raggedright\arraybackslash}p{(\columnwidth - 8\tabcolsep) * \real{0.9301} + 6\tabcolsep}@{}}{%
\emph{\textbf{Input: Game Level}}} \\
&
\multicolumn{4}{>{\raggedright\arraybackslash}p{(\columnwidth - 8\tabcolsep) * \real{0.9301} + 6\tabcolsep}@{}}{%
\emph{\textbf{Output:} - User interactions with a dynamically shuffled
grid.}

\emph{- Completion of the Ultimate Test.}

\emph{- Facial image capture prompt.}

\emph{- Facial image for analysis using Mod 2D CNN.}

\emph{- Result feedback based on Mod 2D CNN analysis.}} \\
&
\multicolumn{4}{>{\raggedright\arraybackslash}p{(\columnwidth - 8\tabcolsep) * \real{0.9301} + 6\tabcolsep}@{}}{%
\emph{\textbf{Procedure:}}

\emph{\textbf{Begin:}}} \\
& &
\multicolumn{3}{>{\raggedright\arraybackslash}p{(\columnwidth - 8\tabcolsep) * \real{0.8652} + 4\tabcolsep}@{}}{%
\emph{\textbf{Enhance game elements:} enhanceGame()}} \\
& &
\multicolumn{3}{>{\raggedright\arraybackslash}p{(\columnwidth - 8\tabcolsep) * \real{0.8652} + 4\tabcolsep}@{}}{%
\emph{\textbf{Dynamically shuffle grid:} shuffleGrid()}} \\
& &
\multicolumn{3}{>{\raggedright\arraybackslash}p{(\columnwidth - 8\tabcolsep) * \real{0.8652} + 4\tabcolsep}@{}}{%
\emph{\textbf{Start the Ultimate Test:} startUltimateTest()}} \\
& &
\multicolumn{3}{>{\raggedright\arraybackslash}p{(\columnwidth - 8\tabcolsep) * \real{0.8652} + 4\tabcolsep}@{}}{%
\emph{\textbf{Set level-specific parameters:}}} \\
& & &
\multicolumn{2}{>{\raggedright\arraybackslash}p{(\columnwidth - 8\tabcolsep) * \real{0.7953} + 2\tabcolsep}@{}}{%
\emph{clickThreshold = 70}} \\
& & &
\multicolumn{2}{>{\raggedright\arraybackslash}p{(\columnwidth - 8\tabcolsep) * \real{0.7953} + 2\tabcolsep}@{}}{%
\emph{timerDuration = 5 minutes}} \\
& & &
\multicolumn{2}{>{\raggedright\arraybackslash}p{(\columnwidth - 8\tabcolsep) * \real{0.7953} + 2\tabcolsep}@{}}{%
\emph{showTimerDuration =10 seconds}} \\
& &
\multicolumn{3}{>{\raggedright\arraybackslash}p{(\columnwidth - 8\tabcolsep) * \real{0.8652} + 4\tabcolsep}@{}}{%
\emph{\textbf{Show timer for memorization:}
showTimer(showTimerDuration)}} \\
& &
\multicolumn{3}{>{\raggedright\arraybackslash}p{(\columnwidth - 8\tabcolsep) * \real{0.8652} + 4\tabcolsep}@{}}{%
\emph{\textbf{while countdownInProgress(timerDuration) do} // Countdown
timer loop}} \\
& & &
\multicolumn{2}{>{\raggedright\arraybackslash}p{(\columnwidth - 8\tabcolsep) * \real{0.7953} + 2\tabcolsep}@{}}{%
\emph{\textbf{Allow user to interact with the grid:}
handleGridInteraction()}} \\
& & &
\multicolumn{2}{>{\raggedright\arraybackslash}p{(\columnwidth - 8\tabcolsep) * \real{0.7953} + 2\tabcolsep}@{}}{%
\emph{\textbf{if allMatchesCompleted() then} // Check for completed
matches}} \\
& & & & \emph{displayAcknowledgment()} \\
& & & & \emph{exitCountdownLoop()} \\
& & &
\multicolumn{2}{>{\raggedright\arraybackslash}p{(\columnwidth - 8\tabcolsep) * \real{0.7953} + 2\tabcolsep}@{}}{%
\emph{\textbf{end if}}} \\
& & &
\multicolumn{2}{>{\raggedright\arraybackslash}p{(\columnwidth - 8\tabcolsep) * \real{0.7953} + 2\tabcolsep}@{}}{%
\emph{\textbf{if exceedClickThreshold(clickThreshold) then} // Check for
click threshold}} \\
& & & & \emph{initiateFacialImageCapture()} \\
& & & & \emph{captureFacialImage()} \\
& & & & \emph{analyzeFacialImageWith2DCNN() // Analyze facial image
using Mod-2D-CNN} \\
& & & & \emph{displayResultFeedback() // Display result feedback based
on Mod-2D-CNN analysis} \\
& & & & \emph{exitCountdownLoop()} \\
& & &
\multicolumn{2}{>{\raggedright\arraybackslash}p{(\columnwidth - 8\tabcolsep) * \real{0.7953} + 2\tabcolsep}@{}}{%
\emph{\textbf{end if}}} \\
& &
\multicolumn{3}{>{\raggedright\arraybackslash}p{(\columnwidth - 8\tabcolsep) * \real{0.8652} + 4\tabcolsep}@{}}{%
\emph{\textbf{end while}}} \\
&
\multicolumn{4}{>{\raggedright\arraybackslash}p{(\columnwidth - 8\tabcolsep) * \real{0.9301} + 6\tabcolsep}@{}}{%
\emph{\textbf{end}}} \\
\bottomrule()
\end{longtable}

\textbf{2.3 Proposed Model}

\textbf{2.3.1 Convolutional Neural Network (CNN)}

Convolutional Neural Networks (CNNs) are a crucial subset of deep
learning, enhancing the analysis of images and videos. They excel at
recognizing spatial patterns within visual data and performing
hierarchical feature extraction, from basic edge detection to complex
textures and object components {[}25-26{]}. CNNs use convolutional
layers with filters and kernels to scan through input images or feature
maps, identifying spatial patterns {[}27{]}. The multi-layered
architecture of CNNs enables a hierarchical learning process,
progressively building upon previously extracted features and improving
accuracy in tasks like classification and object detection {[}28{]}. In
this work, we have proposed a modified CNN (MOD-CNN) algorithm for
dementia prediction tasks. This MOD-CNN is divided into two phases.

1. MOD-1D-CNN is working on the first phase of the game.

2. MOD-2D-CNN is working on the second phase of the game.

\textbf{2.3.2 MOD-1D-CNN in Game level-1 for Health Dataset}

The one-dimensional Convolutional Neural Network, or MOD-1D-CNN, is a
customized model made for processing input sequentially. It is handy for
applications like one-dimensional feature representations and
time-series analysis {[}29-30{]}. Convolutional layers operating across
the temporal axis are incorporated into the design, which enables the
model to recognize patterns and relationships in sequential data.
MOD-1D-CNN uses convolutional processes to automatically learn
hierarchical features, which allows it to recognize pertinent patterns
in sequential input {[}31{]}. This type of CNN is typically used in
cases where the data shows sequential linkages. This technique works
especially well in applications where data is naturally sequential, such
as signal processing and natural language analysis {[}32-34{]}. The
proposed MOD-1D-CNN model that is used to predict the status of dementia
from the health metrics data is a painstakingly constructed system that
uses complex layers to extract extensive features and recognize
patterns. Using Conv1D layers with a kernel size 2x2, accompanied by
ReLU activation function. These first layers serve as feature
extractors, identifying complex patterns from the health metrics data in
sequence. After the Conv1D layers, Dropout layers are placed
strategically to help keep the network from overfitting by randomly
deactivating neurons during training, preventing the network from
depending too much on any one attribute. The next layer is MaxPooling1D,
which efficiently decreases the spatial dimensions while maintaining
important characteristics, improving computing efficiency without
sacrificing essential information. After that, a flattened layer is
placed so that the model reconfigures the data, preparing it for fully
connected layers. By following that, dense layers are positioned for
ReLU activation. This layer's depth facilitates intricate
transformations within the data, enhancing the network's ability to
discern complex patterns. The final layer, which is also a dense layer,
is situated with the sigmoid activation function, facilitating the
critical binary classification between demented and non-demented
individuals. By employing the binary cross-entropy loss function and
Adam optimization with a learning rate of 0.001, this MOD-1D-CNN model
is rigorously compiled for robust training, ensuring accurate and
precise prediction of dementia status from health metrics data. Table 2
shows the detailed network architecture of the MOD-1D-CNN model.
\begin{center}

\textbf{Table 2:} Network Architecture with Number of Parameters of
MOD-1D-CNN Model
\end{center}

\begin{longtable}[]{@{}
  >{\raggedright\arraybackslash}p{(\columnwidth - 8\tabcolsep) * \real{0.1999}}
  >{\raggedright\arraybackslash}p{(\columnwidth - 8\tabcolsep) * \real{0.2000}}
  >{\raggedright\arraybackslash}p{(\columnwidth - 8\tabcolsep) * \real{0.2000}}
  >{\raggedright\arraybackslash}p{(\columnwidth - 8\tabcolsep) * \real{0.2000}}
  >{\raggedright\arraybackslash}p{(\columnwidth - 8\tabcolsep) * \real{0.2000}}@{}}
\toprule()
\begin{minipage}[b]{\linewidth}\raggedright
\textbf{Layer Type}
\end{minipage} & \begin{minipage}[b]{\linewidth}\raggedright
\textbf{Kernel Size}
\end{minipage} & \begin{minipage}[b]{\linewidth}\raggedright
\textbf{Activation}
\end{minipage} & \begin{minipage}[b]{\linewidth}\raggedright
\textbf{Output Shape}
\end{minipage} & \begin{minipage}[b]{\linewidth}\raggedright
\textbf{Number of Parameters}
\end{minipage} \\
\midrule()
\endhead
Conv1D & 2x2 & ReLU & (None, 5, 64) & 192 \\
Conv1D & 2x2 & ReLU & (None, 4, 64) & 8256 \\
Dropout & - & - & (None, 4, 64) & 0 \\
MaxPooling1D & - & - & (None, 2, 64) & 0 \\
Flatten & - & - & (None, 128) & 0 \\
Dense & - & ReLU & (None, 100) & 12900 \\
Dense & - & Sigmoid & (None, 2) & 200 \\
\bottomrule()
\end{longtable}

In summary, the MOD-1D-CNN model architecture effectively leverages
Conv1D layers with ReLU activation,

and MaxPooling1D to extract essential features from the input data
efficiently. The model's ability to learn and generalize from the data
is enhanced by Dropout layers, preventing overfitting issues generated
due to the dataset at a defined rate for effective handling. These
architectural choices collectively contribute to a powerful binary
classification model that can accurately distinguish between demented
and non-demented cases as per the input features of health metrics data
(blood pressure, body temperature, etc.). The training flowchart for the
MOD-1D-CNN model is shown in Fig. 3.

\begin{center}

\includegraphics[width=5.65781in,height=5.05823in]{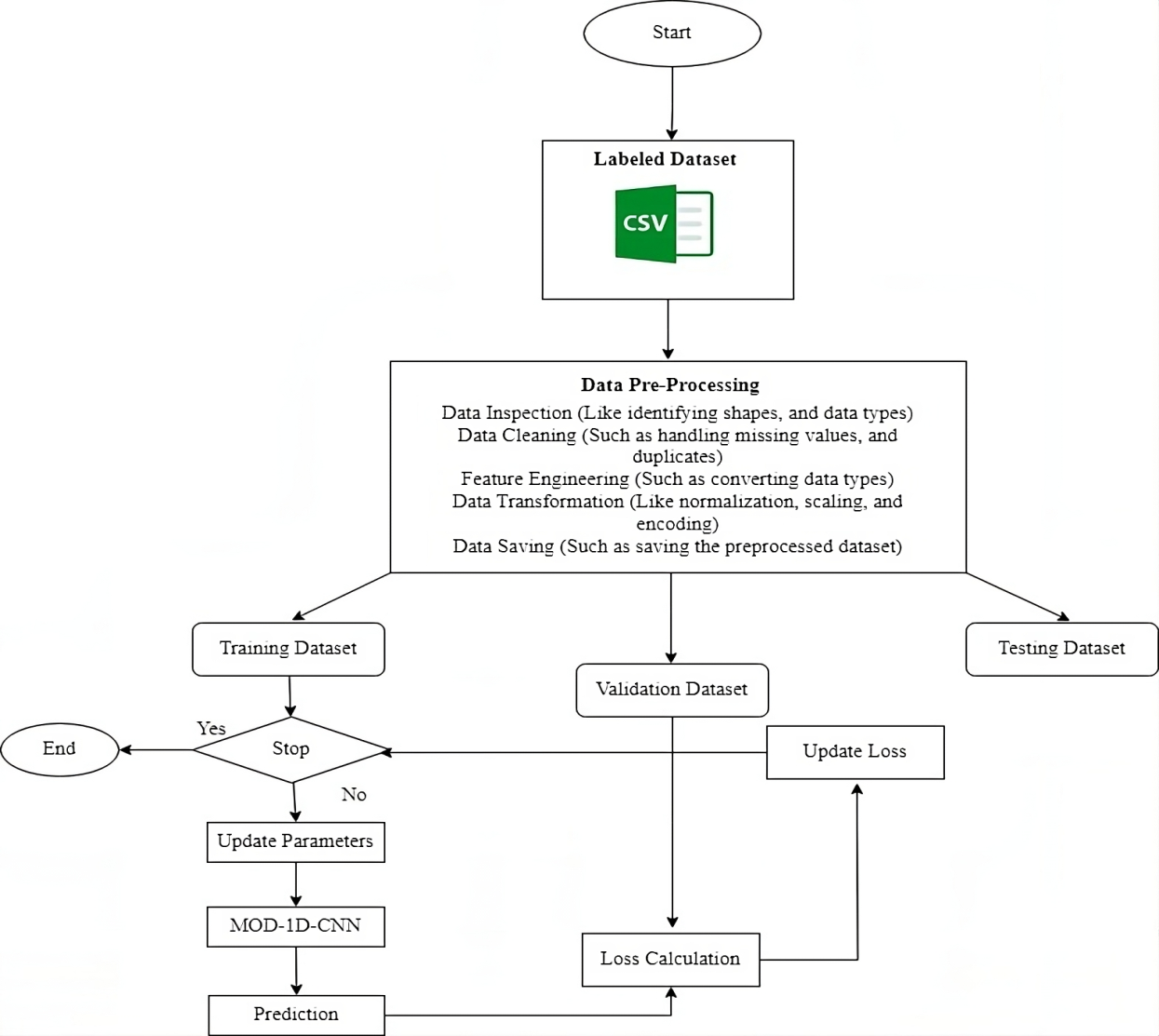}

\textbf{Fig. 3} Training Flowchart of the Proposed MOD-1D-CNN Model
\end{center}

\textbf{2.3.2.1 MOD-1D-CNN Algorithm}

The proposed algorithm for the MOD-1D-CNN model for the classification
of demented and non-demented individuals is shown below:

\begin{longtable}[]{@{}
  >{\raggedright\arraybackslash}p{(\columnwidth - 8\tabcolsep) * \real{0.0647}}
  >{\raggedright\arraybackslash}p{(\columnwidth - 8\tabcolsep) * \real{0.0583}}
  >{\raggedright\arraybackslash}p{(\columnwidth - 8\tabcolsep) * \real{0.0631}}
  >{\raggedright\arraybackslash}p{(\columnwidth - 8\tabcolsep) * \real{0.0518}}
  >{\raggedright\arraybackslash}p{(\columnwidth - 8\tabcolsep) * \real{0.7621}}@{}}
\toprule()
\multicolumn{5}{@{}>{\raggedright\arraybackslash}p{(\columnwidth - 8\tabcolsep) * \real{1.0000} + 8\tabcolsep}@{}}{%
\begin{minipage}[b]{\linewidth}\raggedright
\textbf{ALGORITHM-3: HEALTH PARAMETERS-BASED DEMENTIA PREDICTION USING
MOD-1D-CNN}
\end{minipage}} \\
\midrule()
\endhead
&
\multicolumn{4}{>{\raggedright\arraybackslash}p{(\columnwidth - 8\tabcolsep) * \real{0.9353} + 6\tabcolsep}@{}}{%
\begin{minipage}[t]{\linewidth}\raggedright
\begin{quote}
\emph{\textbf{Input:} - Diabetic status (0 for absence or 1 for
presence)}

\emph{- Blood oxygen level (1(\textless95\%) or 2(95\%-100\%) or
3(\textgreater100\%))}

\emph{- Body temperature: 1(\textless36.5\textsuperscript{o}C) or
2(36.5\textsuperscript{o}C -- 37.5\textsuperscript{o}C) or
3(\textgreater37.5\textsuperscript{o}C)}

\emph{- Heart rate: 1(\textless60bpm) or 2(60bpm -- 100bpm) or
3(\textgreater100bpm)}

\emph{- Weight: 1(\textless50kg) or 2(50kg -- 70kg) or
3(\textgreater70kg)}

\emph{- Age: 1(40-64) or 2(65-70) or 3(75-90)}
\end{quote}
\end{minipage}} \\
&
\multicolumn{4}{>{\raggedright\arraybackslash}p{(\columnwidth - 8\tabcolsep) * \real{0.9353} + 6\tabcolsep}@{}}{%
\begin{minipage}[t]{\linewidth}\raggedright
\begin{quote}
\emph{\textbf{Output:} Prediction result indicating whether the
individual is demented or non-demented.}
\end{quote}
\end{minipage}} \\
&
\multicolumn{4}{>{\raggedright\arraybackslash}p{(\columnwidth - 8\tabcolsep) * \real{0.9353} + 6\tabcolsep}@{}}{%
\begin{minipage}[t]{\linewidth}\raggedright
\begin{quote}
\emph{\textbf{Procedure:}}

\emph{\textbf{Begin}}
\end{quote}
\end{minipage}} \\
& &
\multicolumn{3}{>{\raggedright\arraybackslash}p{(\columnwidth - 8\tabcolsep) * \real{0.8770} + 4\tabcolsep}@{}}{%
\begin{minipage}[t]{\linewidth}\raggedright
\begin{quote}
\emph{\textbf{Initialization:}}
\end{quote}
\end{minipage}} \\
& & &
\multicolumn{2}{>{\raggedright\arraybackslash}p{(\columnwidth - 8\tabcolsep) * \real{0.8139} + 2\tabcolsep}@{}}{%
\begin{minipage}[t]{\linewidth}\raggedright
\begin{quote}
\emph{\textbf{Import necessary libraries:} numpy, keras.}
\end{quote}
\end{minipage}} \\
& & &
\multicolumn{2}{>{\raggedright\arraybackslash}p{(\columnwidth - 8\tabcolsep) * \real{0.8139} + 2\tabcolsep}@{}}{%
\begin{minipage}[t]{\linewidth}\raggedright
\begin{quote}
\emph{\textbf{Load the pre-trained MOD-1D-CNN model:}
load\_1d\_cnn\_model(`mod\_1d\_cnn\_model.h5').}
\end{quote}
\end{minipage}} \\
& &
\multicolumn{3}{>{\raggedright\arraybackslash}p{(\columnwidth - 8\tabcolsep) * \real{0.8770} + 4\tabcolsep}@{}}{%
\begin{minipage}[t]{\linewidth}\raggedright
\begin{quote}
\emph{\textbf{Input Acquisition:}}
\end{quote}
\end{minipage}} \\
& & &
\multicolumn{2}{>{\raggedright\arraybackslash}p{(\columnwidth - 8\tabcolsep) * \real{0.8139} + 2\tabcolsep}@{}}{%
\begin{minipage}[t]{\linewidth}\raggedright
\begin{quote}
\emph{\textbf{takeInput(): Diabetic:} 0 or 1}

\emph{\textbf{takeInput(): Blood Oxygen Level:} 1 or 2 or 3}

\emph{\textbf{takeInput(): Body Temperature:} 1 or 2 or 3}

\emph{\textbf{takeInput(): Heart Rate:} 1 or 2 or 3}

\emph{\textbf{takeInput(): Weight:} 1 or 2 or 3}

\emph{\textbf{takeInput(): Age:} 1 or 2 or 3}
\end{quote}
\end{minipage}} \\
& &
\multicolumn{3}{>{\raggedright\arraybackslash}p{(\columnwidth - 8\tabcolsep) * \real{0.8770} + 4\tabcolsep}@{}}{%
\begin{minipage}[t]{\linewidth}\raggedright
\begin{quote}
\emph{\textbf{Data Preparation:}}
\end{quote}
\end{minipage}} \\
& & &
\multicolumn{2}{>{\raggedright\arraybackslash}p{(\columnwidth - 8\tabcolsep) * \real{0.8139} + 2\tabcolsep}@{}}{%
\emph{\textbf{constructAnArray():}
prepare\_health\_params\_data(diabetic, blood\_oxygen, body\_temp,
heart\_rate, weight, age).}} \\
& & &
\multicolumn{2}{>{\raggedright\arraybackslash}p{(\columnwidth - 8\tabcolsep) * \real{0.8139} + 2\tabcolsep}@{}}{%
\emph{\textbf{normalizeData():}
normalize\_health\_params\_data(health\_params\_data).}} \\
& &
\multicolumn{3}{>{\raggedright\arraybackslash}p{(\columnwidth - 8\tabcolsep) * \real{0.8770} + 4\tabcolsep}@{}}{%
\begin{minipage}[t]{\linewidth}\raggedright
\begin{quote}
\emph{\textbf{Making Predictions:}}
\end{quote}
\end{minipage}} \\
& & &
\multicolumn{2}{>{\raggedright\arraybackslash}p{(\columnwidth - 8\tabcolsep) * \real{0.8139} + 2\tabcolsep}@{}}{%
\emph{\textbf{Call a function for making predictions using the loaded
model:} predict\_dementia\_probability(loaded\_model,
normalized\_health\_params\_data).}} \\
& &
\multicolumn{3}{>{\raggedright\arraybackslash}p{(\columnwidth - 8\tabcolsep) * \real{0.8770} + 4\tabcolsep}@{}}{%
\begin{minipage}[t]{\linewidth}\raggedright
\begin{quote}
\emph{\textbf{Result:}}
\end{quote}
\end{minipage}} \\
& & &
\multicolumn{2}{>{\raggedright\arraybackslash}p{(\columnwidth - 8\tabcolsep) * \real{0.8139} + 2\tabcolsep}@{}}{%
\emph{\textbf{if prediction\_score \textgreater{} 0.5:}}} \\
& & & & \emph{\textbf{Output:} display\_result(``Prediction:
Demented'').} \\
& & &
\multicolumn{2}{>{\raggedright\arraybackslash}p{(\columnwidth - 8\tabcolsep) * \real{0.8139} + 2\tabcolsep}@{}}{%
\emph{\textbf{else:}}} \\
& & & & \emph{\textbf{Output}: display\_result(``Prediction:
Non-Demented'').} \\
& &
\multicolumn{3}{>{\raggedright\arraybackslash}p{(\columnwidth - 8\tabcolsep) * \real{0.8770} + 4\tabcolsep}@{}}{%
\begin{minipage}[t]{\linewidth}\raggedright
\begin{quote}
\emph{\textbf{Termination:}}
\end{quote}
\end{minipage}} \\
& & & & \emph{\textbf{end}} \\
&
\multicolumn{4}{>{\raggedright\arraybackslash}p{(\columnwidth - 8\tabcolsep) * \real{0.9353} + 6\tabcolsep}@{}}{%
\begin{minipage}[t]{\linewidth}\raggedright
\begin{quote}
\emph{\textbf{end}}
\end{quote}
\end{minipage}} \\
\bottomrule()
\end{longtable}

\textbf{2.3.2.2 Working in Game Level 1}

In the first phase (Game Level 1), the typical health data (age, blood
pressure, etc.) from the user are collected and fed into our pre-trained
MOD-1D-CNN algorithm. In this stage, essential health data from the
user, encompass parameters such as age, blood pressure, and other
relevant metrics. It is our initial stage of assessing dementia status
of the prediction assessment. This crucial information serves as the
input for our pre-trained MOD-1D-CNN algorithm, designed specifically
for the analysis of one-dimensional health data. Users in Game Level 1
provide age, blood pressure, and other health metrics, which are
integrated into the MOD-1D-CNN algorithm to extract patterns and
correlations from a one-dimensional health dataset. The MOD-1D-CNN
algorithm, using deep convolutional neural network architecture,
efficiently processes health data to identify subtle relationships
indicating potential health conditions, facilitating comprehensive
health assessments and hence results for the prediction of dementia
status whether the individual is demented or non-demented. Streamlit is
chosen for Game Level 1 hosting to improve user experience. Its
intuitive interface allows easy input of health data and application
stages. Combining MOD-1D-CNN's power with Streamlit's accessibility aims
for a robust, user-centric experience, enhancing informed health
assessment. The working in-game level 1 for the MOD-1D-CNN model is
shown in Fig. 4.

\begin{center}

\includegraphics[width=5.96593in,height=4.78814in]{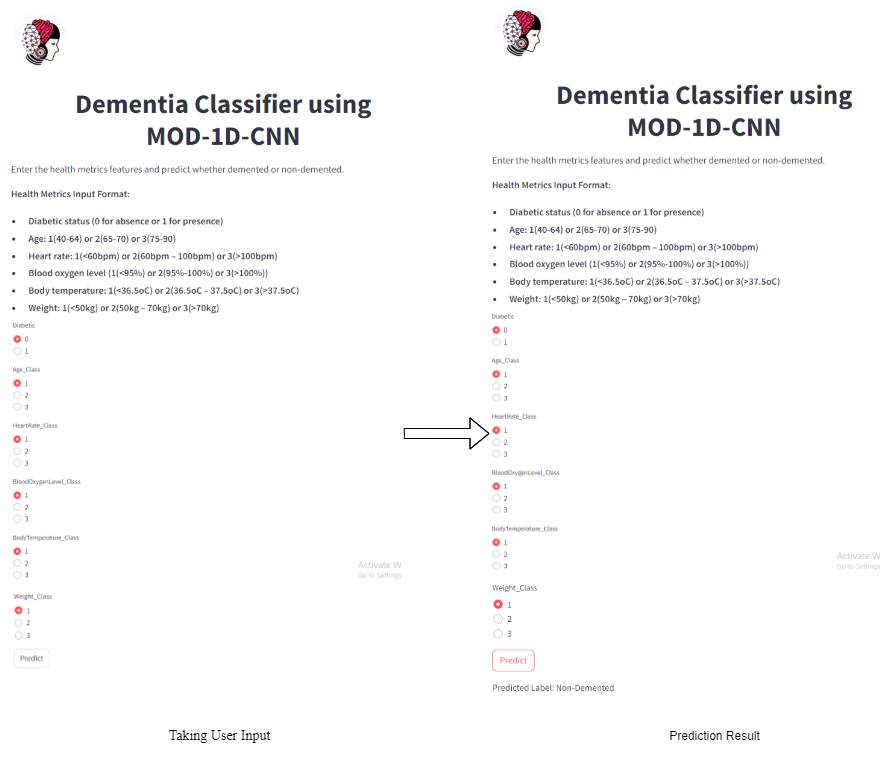}

\textbf{Fig. 4} Taking Health Metrics Input from User for Predicting
Dementia Status
\end{center}

\textbf{2.3.3 MOD-2D-CNN in Game level-2 for Image Dataset}

A more sophisticated version of the conventional 2D Convolutional Neural
Network designed for enhanced image recognition and feature extraction
is called the Mod-2D-CNN {[}35{]}. Key characteristics include residual
connections for smoother information flow in deeper architectures,
dilated convolutions to improve the receptive field, and multi-scale
convolutional layers enabling simultaneous collection of fine- and
coarse-grained data {[}36-38{]}. Spatial information is maintained by
adaptive pooling layers, and feature attention algorithms dynamically
modify feature significance. Batch normalization quickens training
convergence while dropout regularization guards against overfitting
{[}39{]}. With its capacity to adapt to different sizes and complex
characteristics for strong visual identification, this model performs
exceptionally well in a wide range of computer vision applications.
Furthermore, the Mod-2D-CNN effectively recognizes global patterns by
utilizing the capability of dilated convolutions to capture long-range
relationships {[}40{]}. Deeper networks may be trained using residual
connections without running the danger of gradients disappearing or
blowing up. Interpretability is improved by feature attention
techniques, which dynamically shift the emphasis to pertinent areas of
the image {[}41-42{]}.

In the proposed MOD-2D-CNN model, the architecture is designed to
efficiently process and extract the facial features from the input
facial images for binary classification. The model contains a series of
Conv2D layers with a kernel size of 3x3, followed by ReLU activation
functions. These layers are crucial for understanding the fundamental
patterns and characteristics present in the images. After every Conv2D
layer, a MaxPooling 2D operation is applied, which reduces the spatial
dimensions of the feature maps and helps to capture important
information while reducing the computational complexity. By
strengthening the translation invariance of the model and preserving
pertinent information, this pooling procedure helps the model be more
resilient to changes in object location and size. In addition to this,
the flattened layer is also added to convert the 3D output into a 1D
feature vector, facilitating the transition to the fully connected
layers. Furthermore, Dropout layers and Dense layers are also employed
in the MOD-2D-CNN architecture. Dropout layers are used after Dense
layers to prevent overfitting. During training, Dropout introduces a
sort of regularization by randomly deactivating a portion of the
neurons, hence decreasing reliance between them. Better generalization
results from this, strengthening the model's accuracy and resilience to
new data. Finally, the output layer consists of a single neuron with a
sigmoidal activation function, which is used for binary classification,
providing the probability of each input belonging to either class
(demented or non-demented). The model is trained with the binary
cross-entropy loss function, which works well for problems with either 0
or 1 outcomes. It is then optimized using the Adam optimizer, which has
a learning rate of 0.0001. Table 3 shows the detailed network
architecture of the MOD-2D-CNN model.

\begin{center}

\textbf{Table 3:} Network Architecture with Number of Parameters of
MOD-2D-CNN Model
\end{center}

\begin{longtable}[]{@{}
  >{\raggedright\arraybackslash}p{(\columnwidth - 8\tabcolsep) * \real{0.2000}}
  >{\raggedright\arraybackslash}p{(\columnwidth - 8\tabcolsep) * \real{0.2000}}
  >{\raggedright\arraybackslash}p{(\columnwidth - 8\tabcolsep) * \real{0.2000}}
  >{\raggedright\arraybackslash}p{(\columnwidth - 8\tabcolsep) * \real{0.2200}}
  >{\raggedright\arraybackslash}p{(\columnwidth - 8\tabcolsep) * \real{0.1800}}@{}}
\toprule()
\begin{minipage}[b]{\linewidth}\raggedright
\textbf{Layer Type}
\end{minipage} & \begin{minipage}[b]{\linewidth}\raggedright
\textbf{Kernel Size}
\end{minipage} & \begin{minipage}[b]{\linewidth}\raggedright
\textbf{Activation}
\end{minipage} & \begin{minipage}[b]{\linewidth}\raggedright
\textbf{Output Shape}
\end{minipage} & \begin{minipage}[b]{\linewidth}\raggedright
\textbf{Number of Parameters}
\end{minipage} \\
\midrule()
\endhead
Cov2D & 3x3 & ReLU & (None, 222, 222, 32) & 896 \\
MaxPooling2D & - & - & (None, 111, 111, 32) & 0 \\
Cov2D\_1 & 3x3 & ReLU & (None, 109, 109, 64) & 14,496 \\
MaxPooling2D\_1 & - & - & (None, 54, 54, 64) & 0 \\
Cov2D\_2 & 3x3 & ReLU & (None, 52, 52, 128) & 73,856 \\
MaxPooling2D\_2 & - & - & (None, 26, 26, 128) & 0 \\
Cov2D\_3 & 3x3 & ReLU & (None, 24, 24, 256) & 2,95, 168 \\
MaxPooling2D\_3 & - & - & (None, 12, 12, 256) & 0 \\
Flatten & - & - & (None, 36,864) & 0 \\
Dense & - & ReLU & (None, 512) & 18,874,880 \\
Dropout & - & - & (None, 512) & 0 \\
Dense\_1 & - & ReLU & (None, 256) & 1,131,328 \\
Dropout\_1 & - & - & (None, 256) & 0 \\
Dense\_2 & - & Sigmoid & (None, 1) & 257 \\
\bottomrule()
\end{longtable}

In summary, the MOD-2D-CNN model architecture successfully extracts
critical features from the input images by utilizing Conv2D, and
MaxPooling layers with ReLU activation. Dropout layers improve the
model's capacity to learn and generalize from the data, avoiding
overfitting. Together, these architectural decisions provide a potent
binary classification model that effectively discriminates between
non-demented facial images and demented facial images. The training
flowchart depicted in Fig. 5 outlines the sequential steps involved in
training the MOD-2D-CNN model. The representation encapsulates the
entire training process, illustrating key stages such as data
preprocessing, data splitting, model initialization, updation of
parameters, and loss calculation at the time of the training process.

\begin{center}

\includegraphics[width=5.04323in,height=6.67007in]{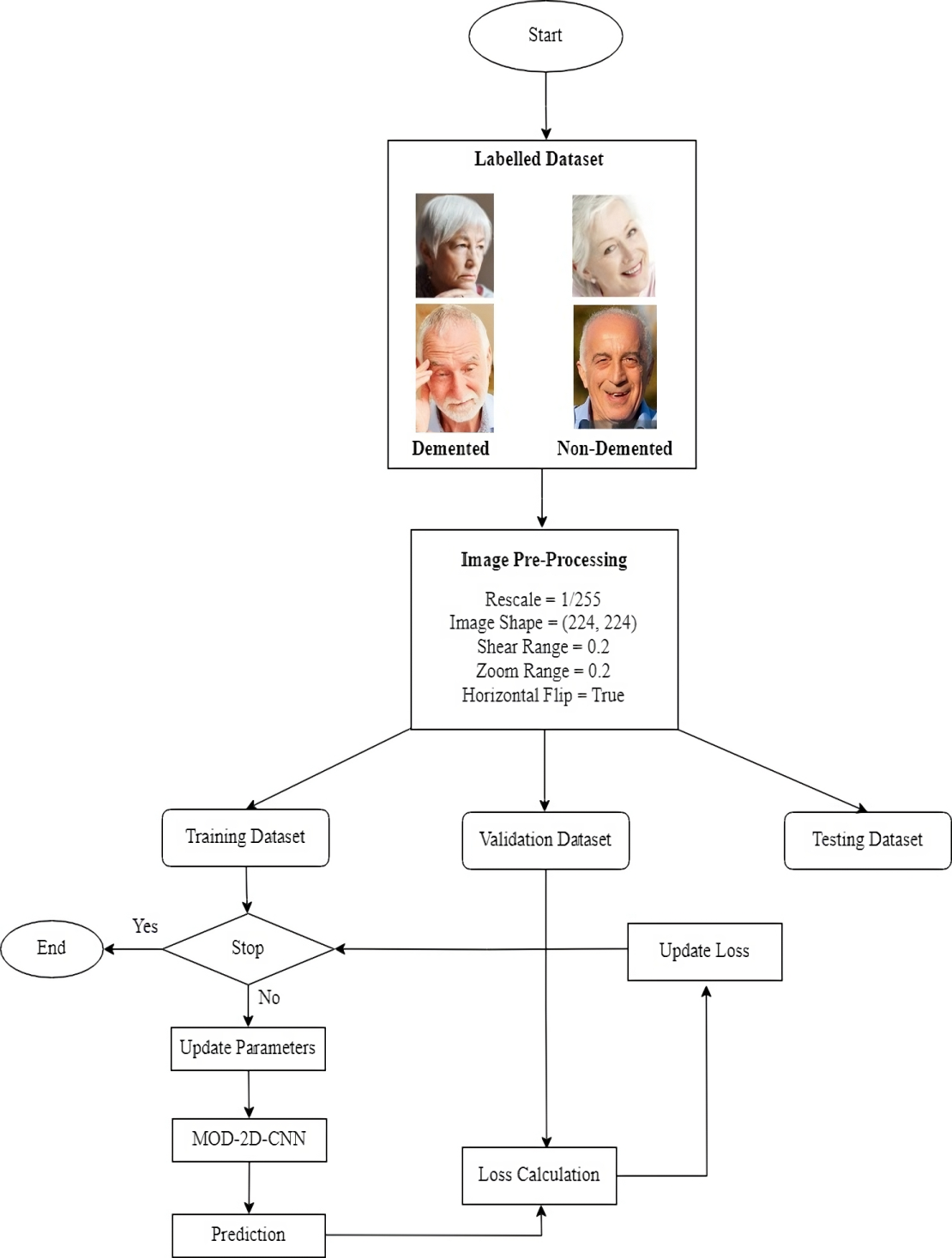}

\textbf{Fig. 5} Training Flowchart of the Proposed MOD-2D-CNN Model
\end{center}

\textbf{2.3.3.1 MOD-2D-CNN Algorithm}

The proposed algorithm for the MOD-2D-CNN model for the classification
of demented and non-demented individuals is shown below,

\begin{longtable}[]{@{}
  >{\raggedright\arraybackslash}p{(\columnwidth - 8\tabcolsep) * \real{0.0592}}
  >{\raggedright\arraybackslash}p{(\columnwidth - 8\tabcolsep) * \real{0.0654}}
  >{\raggedright\arraybackslash}p{(\columnwidth - 8\tabcolsep) * \real{0.0654}}
  >{\raggedright\arraybackslash}p{(\columnwidth - 8\tabcolsep) * \real{0.0592}}
  >{\raggedright\arraybackslash}p{(\columnwidth - 8\tabcolsep) * \real{0.7508}}@{}}
\toprule()
\multicolumn{5}{@{}>{\raggedright\arraybackslash}p{(\columnwidth - 8\tabcolsep) * \real{1.0000} + 8\tabcolsep}@{}}{%
\begin{minipage}[b]{\linewidth}\raggedright
\textbf{ALGORITHM-4: FACIAL CLASSIFICATION FOR DEMENTIA DETECTION USING
MOD-2D-CNN MODEL}
\end{minipage}} \\
\midrule()
\endhead
&
\multicolumn{4}{>{\raggedright\arraybackslash}p{(\columnwidth - 8\tabcolsep) * \real{0.9408} + 6\tabcolsep}@{}}{%
\emph{\textbf{Input:} Facial image (live capture or pre-loaded)}} \\
&
\multicolumn{4}{>{\raggedright\arraybackslash}p{(\columnwidth - 8\tabcolsep) * \real{0.9408} + 6\tabcolsep}@{}}{%
\emph{\textbf{Output:} Prediction result indicating whether the
individual is demented or non-demented using facial analysis.}} \\
&
\multicolumn{4}{>{\raggedright\arraybackslash}p{(\columnwidth - 8\tabcolsep) * \real{0.9408} + 6\tabcolsep}@{}}{%
\emph{\textbf{Procedure:}}

\emph{\textbf{begin}}} \\
& &
\multicolumn{3}{>{\raggedright\arraybackslash}p{(\columnwidth - 8\tabcolsep) * \real{0.8754} + 4\tabcolsep}@{}}{%
\emph{\textbf{Initialization:}}} \\
& & &
\multicolumn{2}{>{\raggedright\arraybackslash}p{(\columnwidth - 8\tabcolsep) * \real{0.8100} + 2\tabcolsep}@{}}{%
\emph{\textbf{Import necessary libraries:} os, cv2, numpy, keras.}} \\
& & &
\multicolumn{2}{>{\raggedright\arraybackslash}p{(\columnwidth - 8\tabcolsep) * \real{0.8100} + 2\tabcolsep}@{}}{%
\emph{\textbf{Load the pre-trained model:}
keras.models.load\_model(`mod\_2d\_cnn\_model.h5')}} \\
& &
\multicolumn{3}{>{\raggedright\arraybackslash}p{(\columnwidth - 8\tabcolsep) * \real{0.8754} + 4\tabcolsep}@{}}{%
\emph{\textbf{Preprocessing Steps:}}} \\
& & &
\multicolumn{2}{>{\raggedright\arraybackslash}p{(\columnwidth - 8\tabcolsep) * \real{0.8100} + 2\tabcolsep}@{}}{%
\emph{\textbf{if inputimage is liveCapture:}}} \\
& & & & \emph{\textbf{use a camera to capture a live facial image:}
capturedImageFromCamera()} \\
& & &
\multicolumn{2}{>{\raggedright\arraybackslash}p{(\columnwidth - 8\tabcolsep) * \real{0.8100} + 2\tabcolsep}@{}}{%
\emph{\textbf{else:}}} \\
& & & & \emph{\textbf{load the image from the specified file path:}
loadImageFromFile(inputImage)} \\
& & &
\multicolumn{2}{>{\raggedright\arraybackslash}p{(\columnwidth - 8\tabcolsep) * \real{0.8100} + 2\tabcolsep}@{}}{%
\emph{\textbf{end if}}} \\
& & &
\multicolumn{2}{>{\raggedright\arraybackslash}p{(\columnwidth - 8\tabcolsep) * \real{0.8100} + 2\tabcolsep}@{}}{%
\emph{\textbf{convert image to array:}
convertImageToArray(capturedImage)}} \\
& & &
\multicolumn{2}{>{\raggedright\arraybackslash}p{(\columnwidth - 8\tabcolsep) * \real{0.8100} + 2\tabcolsep}@{}}{%
\emph{\textbf{expand dimensions to match the expected input shape:}
expandDimensions(imageArray)}} \\
& & &
\multicolumn{2}{>{\raggedright\arraybackslash}p{(\columnwidth - 8\tabcolsep) * \real{0.8100} + 2\tabcolsep}@{}}{%
\emph{\textbf{normalize the pixel values:}
normalizePixels(expandedImageArray)}} \\
& &
\multicolumn{3}{>{\raggedright\arraybackslash}p{(\columnwidth - 8\tabcolsep) * \real{0.8754} + 4\tabcolsep}@{}}{%
\emph{\textbf{Making Predictions:}}} \\
& & &
\multicolumn{2}{>{\raggedright\arraybackslash}p{(\columnwidth - 8\tabcolsep) * \real{0.8100} + 2\tabcolsep}@{}}{%
\emph{\textbf{making predictions using the loaded model:}
loadedModel.predict(preprocessedImage)}} \\
& & & & \emph{\textbf{function input:} Preprocessed image.} \\
& & & & \emph{\textbf{function output:} Prediction score (probability of
being demented).} \\
& &
\multicolumn{3}{>{\raggedright\arraybackslash}p{(\columnwidth - 8\tabcolsep) * \real{0.8754} + 4\tabcolsep}@{}}{%
\emph{\textbf{Result:}}} \\
& & &
\multicolumn{2}{>{\raggedright\arraybackslash}p{(\columnwidth - 8\tabcolsep) * \real{0.8100} + 2\tabcolsep}@{}}{%
\emph{\textbf{if predictionscore \textgreater{} 0.5:}}} \\
& & & & \emph{\textbf{output:} ``Prediction: Demented.''} \\
& & &
\multicolumn{2}{>{\raggedright\arraybackslash}p{(\columnwidth - 8\tabcolsep) * \real{0.8100} + 2\tabcolsep}@{}}{%
\emph{\textbf{else:}}} \\
& & & & \emph{\textbf{output:} ``Prediction: Non-Demented.''} \\
& & &
\multicolumn{2}{>{\raggedright\arraybackslash}p{(\columnwidth - 8\tabcolsep) * \real{0.8100} + 2\tabcolsep}@{}}{%
\emph{\textbf{end if}}} \\
& &
\multicolumn{3}{>{\raggedright\arraybackslash}p{(\columnwidth - 8\tabcolsep) * \real{0.8754} + 4\tabcolsep}@{}}{%
\emph{\textbf{Termination:}}} \\
& & &
\multicolumn{2}{>{\raggedright\arraybackslash}p{(\columnwidth - 8\tabcolsep) * \real{0.8100} + 2\tabcolsep}@{}}{%
\emph{\textbf{end}}} \\
&
\multicolumn{4}{>{\raggedright\arraybackslash}p{(\columnwidth - 8\tabcolsep) * \real{0.9408} + 6\tabcolsep}@{}}{%
\emph{\textbf{end}}} \\
\bottomrule()
\end{longtable}

\textbf{2.3.3.2 Working in Game Level 2}

In the second phase (Game Level 2), we collect the facial image (live or
pre-loaded) from the user and feed it into our pre-trained MOD-2D-CNN
algorithm. During this stage, users are prompted to provide either a
live or pre-loaded facial image. It's our second level of examination
for the user to predict the dementia status. This image is then
seamlessly integrated into our advanced MOD-2D-CNN (Modified 2D
Convolutional Neural Network) algorithm, which has been pre-trained to
discern crucial patterns and features indicative of dementia. Our
cutting-edge MOD-2D-CNN algorithm serves as the backbone for predicting
dementia status, categorizing users into either the demented or
non-demented group based on the intricate analysis of facial
characteristics. Streamlit is chosen for its user-friendly interface,
enhancing user interaction and accessibility in our dementia prediction
tool. It simplifies the process of submitting facial images and
providing prompt results on dementia status. The working in-game level 2
for the MOD-2D-CNN model is shown in Fig. 6 and Fig. 7 for both live
capture and pre-loaded.

\begin{center}

\includegraphics[width=5.35754in,height=3.05026in]{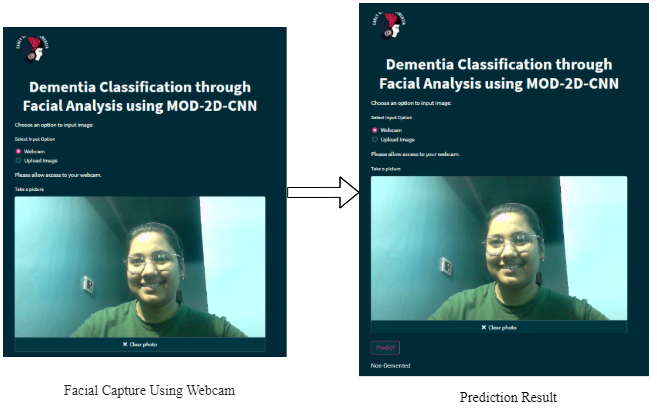}

\textbf{Fig. 6} Facial Image Capture from User Using Live Camera

\includegraphics[width=4.34371in,height=3.34477in]{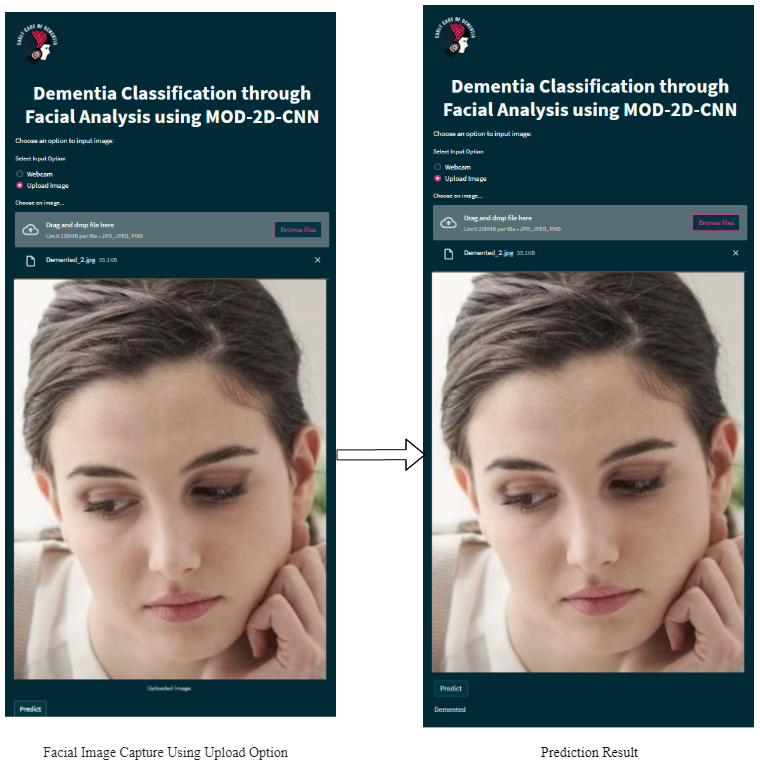}

\textbf{Fig. 7} Facial Image Capture from User using Upload Option
(Pre-loaded)
\end{center}

\textbf{2.3.4 Rule-based Weightage Method for Decision-Making}

Table 4, comprises two columns detailing the weights assigned to two
distinct convolutional neural network models for dementia prediction.
MOD-1D-CNN is allocated a weightage of 30\%, while MOD-2D-CNN takes
precedence with a more substantial weightage of 70\%. The subsequent
rows delineate various scenarios denoted by the decision outcomes,
showcasing the model's predictions and the final decisions based on
their weighted contributions.
\begin{center}

\textbf{Table 4:} Decision-making using rule-based weightage method
\end{center}
\begin{longtable}[]{@{}
  >{\raggedright\arraybackslash}p{(\columnwidth - 4\tabcolsep) * \real{0.2493}}
  >{\raggedright\arraybackslash}p{(\columnwidth - 4\tabcolsep) * \real{0.2332}}
  >{\raggedright\arraybackslash}p{(\columnwidth - 4\tabcolsep) * \real{0.5174}}@{}}
\toprule()
\begin{minipage}[b]{\linewidth}\raggedright
\textbf{MOD-1D-CNN}

\textbf{(30\% weightage)}
\end{minipage} & \begin{minipage}[b]{\linewidth}\raggedright
\textbf{MOD-2D-CNN}

\textbf{(70\% weightage)}
\end{minipage} & \begin{minipage}[b]{\linewidth}\raggedright
\textbf{Decision}
\end{minipage} \\
\midrule()
\endhead
1 & 1 & Demented \\
0 & 1 & Demented with a high probability \\
1 & 0 & Non-Demented with a high probability \\
0 & 0 & Non-Demented \\
\bottomrule()
\end{longtable}

\textbf{3. EXPERIMENTAL SETUP}

\textbf{3.1 Dataset Description for MOD-1D-CNN}

The dataset utilized for training and evaluating the MOD-1D-CNN for
cognitive health assessment comprises 1000 data instances sourced from
the Apollo diagnostic center Kolkata, India. The central target
variable, labeled ``Dementia,'' assumes binary values (0 or 1),
representing the absence (0) or presence (1) of dementia. Additionally,
a binary variable labeled ``Diabetic'' is incorporated, indicating the
absence (0) or presence (1) of diabetes. Alongside these categorical
attributes, the dataset includes five quantitative features, namely
``Age,'' ``Blood Oxygen Level,'' ``Heart Rate,'' ``Body Temperature,''
and ``Weight,'' providing crucial health metrics for the assessment. The
dataset undergoes a standard split into training (80\%) and testing
(20\%) sets to facilitate model training and evaluation. Within the
training set, a further subdivision designates 20\% of the data (160
instances) as a validation subset, utilized during the training process
to enhance model performance and prevent overfitting. Important
Variables and Measures Included in the dataset are represented in Table
5 below.

\begin{center}

\textbf{Table 5:} Dataset Description for MOD-1D-CNN
\end{center}

\begin{longtable}[]{@{}
  >{\raggedright\arraybackslash}p{(\columnwidth - 2\tabcolsep) * \real{0.2401}}
  >{\raggedright\arraybackslash}p{(\columnwidth - 2\tabcolsep) * \real{0.7599}}@{}}
\toprule()
\begin{minipage}[b]{\linewidth}\raggedright
\textbf{Measures}
\end{minipage} & \begin{minipage}[b]{\linewidth}\raggedright
\textbf{Description}
\end{minipage} \\
\midrule()
\endhead
\textbf{Dementia} & Binary variable indicating the presence (1) or
absence (0) of dementia. \\
\textbf{Diabetic} & Binary variable denoting the presence (1) or absence
(0) of diabetes. \\
\textbf{Age} & Quantitative features representing the age of individuals
in the dataset. \\
\textbf{Blood Oxygen Level} & A quantitative measure indicating the
blood oxygen level in the subjects. \\
\textbf{Heart Rate} & Quantitative measure representing the heart rate
of individuals. \\
\textbf{Body Temperature} & A quantitative measure indicating the body
temperature of the subjects. \\
\textbf{Weight} & Quantitative features representing the weight of
individuals. \\
\bottomrule()
\end{longtable}

\textbf{3.2 Dataset Description for MOD-2D-CNN}

The facial image dataset employed for 2D Convolutional Neural Network
(CNN) analysis in this research is meticulously crafted to facilitate
the accurate prediction of dementia based on facial traits. Unlike
publicly available datasets, this proprietary collection is purposefully
designed with precision and care, ensuring its alignment with the
specific requirements of cognitive health assessment. The dataset is
categorized into two distinct classes: demented and non-demented,
reflecting the binary nature of the prediction task. The demented class
comprises 900 facial images, each capturing distinct facial expressions
and features indicative of dementia, while the non-demented class
equally contributes 900 images for a comprehensive evaluation. The
training set consists of 1260 images to get an in-depth understanding of
facial features. The 630 images that each class---demented and
non-demented---represents offer extensive information for training the
2D CNN model. The testing set, which consists of 360 images (180 from
each class), is essential for evaluating the generalization capabilities
of the model. This portion of the dataset is left unaltered during the
model-training procedure, providing an objective assessment of the
model's prediction capabilities on data that hasn't been seen before. A
validation set of 180 images (90 from each class) is used in the model
training phase to adjust the model parameters and avoid overfitting.
This iterative approach guarantees the trained model's resilience and
dependability. Table 6 describes the detailed dataset split ratio for
the demented as well as non--demented class of facial images.

\begin{center}

\textbf{Table 6:} Dataset Split for Dementia Classification using Facial
Images
\end{center}

\begin{longtable}[]{@{}
  >{\raggedright\arraybackslash}p{(\columnwidth - 8\tabcolsep) * \real{0.1999}}
  >{\raggedright\arraybackslash}p{(\columnwidth - 8\tabcolsep) * \real{0.2000}}
  >{\raggedright\arraybackslash}p{(\columnwidth - 8\tabcolsep) * \real{0.2000}}
  >{\raggedright\arraybackslash}p{(\columnwidth - 8\tabcolsep) * \real{0.2000}}
  >{\raggedright\arraybackslash}p{(\columnwidth - 8\tabcolsep) * \real{0.2000}}@{}}
\toprule()
\begin{minipage}[b]{\linewidth}\raggedright
\textbf{Dataset Split}
\end{minipage} & \begin{minipage}[b]{\linewidth}\raggedright
\textbf{Demented Class}
\end{minipage} & \begin{minipage}[b]{\linewidth}\raggedright
\textbf{Non-Demented Class}
\end{minipage} & \begin{minipage}[b]{\linewidth}\raggedright
\textbf{Total Images}
\end{minipage} & \begin{minipage}[b]{\linewidth}\raggedright
\textbf{Reference}
\end{minipage} \\
\midrule()
\endhead
\textbf{Training Set} & 630 & 630 & 1260 & \\
\textbf{Testing Set} & 180 & 180 & 360 & {[}72-75{]} \\
\textbf{Validation Set} & 90 & 90 & 180 & \\
& & & \textbf{Total =} 1800 & \\
\bottomrule()
\end{longtable}

It is crucial to emphasize that this facial image dataset is not sourced
from public repositories but is instead a purpose-built collection
created by the research team. This deliberate approach guarantees the
dataset's tailored relevance to the research objectives, providing a
solid foundation for accurate and insightful dementia prediction through
facial image analysis.

\textbf{3.3 Execution Platform}

The implementation is carried over the Jupyter Notebook Environment
using Python Programming Language for the preparation of models. All the
results are analyzed and conducted on the underlying system
configuration.

\begin{itemize}
\item
  Intel core i5 \(10^{th}\) generation processor.
\item
  12 GB of RAM.
\item
  256 GB SSD.
\end{itemize}

\textbf{3.4 Extraction of Features and Dataset Preprocessing of
MOD-1D-CNN}

In the pursuit of developing a robust cognitive health assessment
MOD-1D-CNN model, the initial phase involves meticulous preprocessing of
the dataset obtained from a diagnostic center. This dataset, rich in
diverse health metrics, is meticulously divided into distinct classes,
such as heart rate, age, body temperature, blood oxygen level, weight,
diabetes, and dementia which is our target variable. These classes serve
as crucial indicators in the evaluation of dementia. In the case of
MOD-1D-CNN, the following preprocessing procedures are applied,

\begin{enumerate}
\def\labelenumi{\alph{enumi}.}
\item
  \emph{Numerical Features Categorization:} The categorization involves
  defining explicit classes for key health metrics, ensuring a
  comprehensive representation of cognitive health. For heart rate,
  three classes are delineated: Class I signifies a heart rate below 60
  bpm, Class II spans heart rates between 60 and 100 bpm, and Class III
  designates heart rates exceeding 100 bpm. Age categories encompass
  Class I (40-64 years), Class II (65-74 years), and Class III (75-90
  years). Body temperature is stratified into Class I (\textless36.5°C),
  Class II (36.5°C-37.5°C), and Class III (\textgreater37.5°C).
  Similarly, blood oxygen levels are categorized as Class I
  (\textless95\%), Class II (95\%-100\%), and Class III
  (\textgreater100\%). Weight classes are defined as Class I
  (\textless50 kg), Class II (50-70 kg), and Class III (\textgreater70
  kg).
\item
  \emph{Data Preprocessing:} Features and target variables are separated
  in this step, where X represents features, and Y represents the target
  variable ``Dementia''.
\item
  \emph{Oversampling using Synthetic Minority Over-sampling Technique
  (SMOTE):} SMOTE (Synthetic Minority Over-sampling Technique) is
  applied to oversample the minority in the training dataset.
\item
  \emph{Normalization/Scaling:} Standard scaling is performed on the
  features to normalize the data using `StandardScaler' to ensure
  uniformity.
\item
  \emph{Batch Processing:} For the MOD-1D-CNN model in our case the
  dataset is divided into 32 batches in size to accelerate the process
  of training so that multiple samples of data inputs can be processed
  at the same time.
\end{enumerate}

\textbf{3.5 Feature Mapping of MOD-1D-CNN}

Feature mapping is a key part of aligning different health parameters
with the architecture of the neural network. This makes it easier to
find patterns and correlations in the input data {[}43-44{]}. In the
context of our research, feature mapping involves the systematic
arrangement and representation of health-related features such as
Diabetic status, Age classification, Heart Rate classification, Blood
Oxygen Level classification, Body Temperature classification, and Weight
classification. These features serve as the input variables for the
subsequent MOD-1D Convolutional Neural Network (CNN) model, which aims
to predict dementia based on the provided health metrics.

\textbf{3.6 Extraction of Features and Dataset Preprocessing of
MOD-2D-CNN}

Feature extraction and dataset preprocessing are important steps in the
MOD-2D-CNN model for getting the facial images of people with and
without dementia ready for binary classification. Feature extraction
entails converting unstructured data---in this case, facial images of
people with and without dementia---into a format that the model can use
{[}45-46{]}. In the case of MOD-2D-CNN, the following preprocessing
techniques are applied:

\begin{enumerate}
\def\labelenumi{\alph{enumi}.}
\item
  \emph{Rescaling:} The pixel values of the images are rescaled by
  dividing them by 255. This step ensures that the pixel values are
  within the range of 0 to 1.
\item
  \emph{Data Augmentation:} Various augmentation techniques are employed
  to increase the variability and diversity of the training dataset.
  These techniques include rotation, height, and width shifting,
  shearing, zooming, and horizontal flipping. Data augmentation helps in
  mitigating overfitting issues and generalizes the model's ability to
  handle unseen data.
\item
  \emph{Target Size:} The images are resized to a target size of 224x224
  pixels. Resizing the images to a fixed size allows for consistent
  input dimensions across the dataset and facilitates an efficient way
  of processing with the neural network architecture.
\item
  \emph{Color Mode:} The color mode of the images is set to RGB,
  representing red, green, and blue color channels. RGB color images are
  well suited for the task of computer vision and provide richer
  information for the model to learn patterns from them.
\item
  \emph{Batch Processing:} The dataset is divided into batches of size
  32 in our case. Batch processing enables parallel computation and
  accelerates the process of training, as multiple images can be
  processed simultaneously.
\item
  \emph{Class Mode:} The class mode is set to ``binary'' to reflect the
  fact that this is a binary classification task using the labels
  ``DEMENTED'' and ``NON-DEMENTED''.
\end{enumerate}

These preprocessing techniques, such as rescaling, data augmentation,
resizing, color mode selection, and batch processing, collectively
enhance the model's ability to learn patterns from the pixel values
associated with the input images and make accurate predictions based on
those learned features.

\textbf{3.7 Feature Mapping of MOD-2D-CNN}

Feature mapping is mainly used to visualize the activation patterns
across different convolutional hidden layers. This process allows us to
understand and interpret the representation of the learned features by
the model {[}47{]}. Feature mapping incorporates processes like
extracting and visualizing the intermediate feature maps produced by
each convolutional layer {[}48{]}. These feature maps identify
particular structures, patterns, or textures in the input images. We can
find more insights about how the model is learning and which features
are being prioritized by viewing these feature maps at every
convolutional layer. An example image is taken from the dataset and
passed through the MOD-2D-CNN model to obtain the layer activations. As
shown in Fig. 8, the activations are then processed and displayed to
provide a grid of images, each of which represents the activation of a
particular filter inside a certain convolutional layer.

\begin{center}

\includegraphics[width=6.27083in,height=0.68229in]{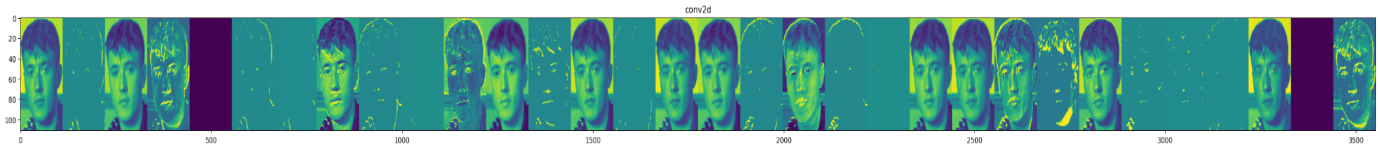}
\end{center}

\begin{center}
  a. First Conv2D layer with 32 filters
\end{center}

\begin{center}
\includegraphics[width=6.27083in,height=0.46163in]{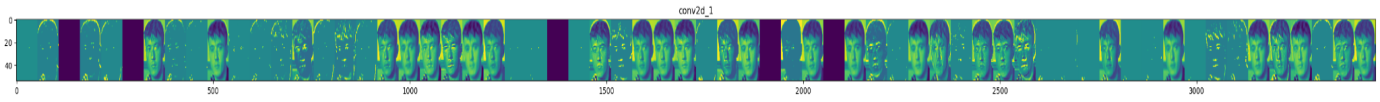}
\end{center}

\begin{center}
  b. Second Conv2D layer with 64 filters
\end{center}

\begin{center}
\includegraphics[width=6.27083in,height=0.56771in]{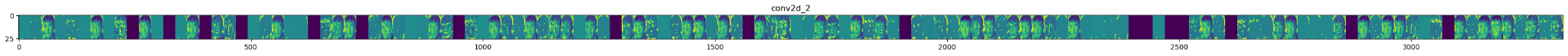}
\end{center}

\begin{center}
  c. Third Conv2D layer with 128 filters
\end{center}

\begin{center}
\includegraphics[width=6.27083in,height=0.73629in]{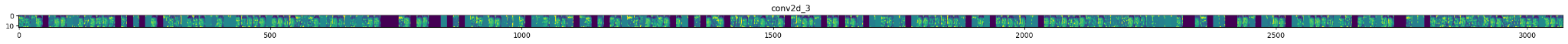}
\end{center}

\begin{center}
  d. Fourth Conv2D layer with 256 filters
\end{center}
\begin{center}

\textbf{Fig. 8} Feature Mapping of the Convolutional Layers

\end{center}

By visualizing the visual interpretations of the feature maps of the
conv2D layers, we can observe that the activations of individual filters
and their spatial relationships are represented. This visualization
explains how the model identifies many traits and patterns that the
model recognizes as the input propagates through the neural network.
Furthermore, it provides insights into the learned representation's
hierarchical structure, in which more abstract and complex features are
captured by higher layers.

\textbf{4. RESULT ANALYSIS}

\textbf{4.1 Evaluation Metrics}

Accuracy, precision, recall, F1-score, and the confusion matrix are some
of the evaluation metrics that are frequently used to find the
effectiveness of the MOD-1D-CNN and MOD-2D-CNN models. These metrics
offer a thorough grasp of the model's classification abilities and its
capability to distinguish between non-demented and demented patients
effectively. In this subsection, we define each metric, present their
mathematical formulae, and discuss their specific uses in evaluating the
performance of the demented detection system. Let \emph{TP} be the truly
predicted demented cases, \emph{TN} be the truly predicted non-demented
cases, \emph{FP} be the falsely predicted demented cases and \emph{FN}
be the falsely predicted non-demented cases.

\begin{enumerate}
\def\labelenumi{\Alph{enumi}.}
\item
  \textbf{Accuracy:}
\end{enumerate}

Accuracy is a pivotal metric that gauges the precision of a
classification model's forecasts, indicating the extent of accurately
classified occasions within the dataset {[}49{]}. A high accuracy score
shows the model's adequacy in foreseeing correct classes {[}50{]}. In
any case, more is needed in imbalanced datasets, because it may distort
the model's execution.

\begin{center}

\(Accuracy = \ \frac{TP\  + \ TN}{TP\  + \ TN\  + \ FP\  + \ FN}\)
\ldots\ldots\ldots{} (1)

\end{center}

Accuracy is commonly used to evaluate the overall performance of the
MOD-1D-CNN and MOD-2D-CNN models and provide a single metric for
classifying the classes in the dataset.

\begin{enumerate}
\def\labelenumi{\Alph{enumi}.}
\setcounter{enumi}{1}
\item
  \textbf{Precision:}
\end{enumerate}

Precision expresses the ratio of true positives to total positive
instances, which indicates how accurate a model is in making positive
predictions {[}51{]}. It measures the model's accuracy in designating
positive examples, guaranteeing their applicability. When false
positives are expensive or unwanted, high accuracy is essential
{[}52{]}.

\begin{center}

\(Precision\  = \ \frac{TP}{TP + \ FP}\) \ldots\ldots\ldots.(2)
\end{center}

When it comes to avoiding false positive mistakes, precision is very
critical. It assists in assessing how well the system detects cases of
dementia.

\begin{enumerate}
\def\labelenumi{\Alph{enumi}.}
\setcounter{enumi}{2}
\item
  \textbf{Recall (Sensitivity):}
\end{enumerate}

The recall measures a model's ability to accurately distinguish all
positive instances in a dataset, indicating its ability to recognize and
capture the most positive instances without missing many, particularly
in situations where false negatives are costly or destructive {[}53{]}.

\begin{center}

\(Recall = \ \frac{TP}{TP + \ FN}\) \ldots\ldots\ldots.(3)
\end{center}

To minimize false negatives and identify as many dementia patients as
feasible, recall is crucial {[}54{]}. It aids in evaluating the
sensitivity of the MOD-1D-CNN and MOD-2D-CNN models in precisely
identifying instances of dementia.

\begin{enumerate}
\def\labelenumi{\Alph{enumi}.}
\setcounter{enumi}{3}
\item
  \textbf{F1-Score:}
\end{enumerate}

The F1-score is a metric that evaluates a classification model's
performance by combining precision and recall. It's useful in uneven
class distributions or when false positives and false negatives have
varying importance {[}55{]}. A higher F1-score indicates a better
balance between these metrics, showcasing an optimal trade-off.

\begin{center}

\(F1 - score\  = \ 2 \times \frac{Precision\  \times Recall}{Precision\  + \ Recall}\)
\ldots\ldots\ldots.(4)

\end{center}

A thorough assessment of the system's performance that strikes a balance
between precision and recall is provided by the F1-score {[}56{]}. When
minimizing false positive and false negative mistakes is necessary, it
is frequently utilized.

\begin{enumerate}
\def\labelenumi{\Alph{enumi}.}
\setcounter{enumi}{4}
\item
  \textbf{ROC - AUC Score:}
\end{enumerate}

The Receiver Operating Characteristic (ROC) curve and ROC-AUC score are
used to evaluate a classification model's performance. The ROC curve
represents the trade-off between the genuine positive rate (sensitivity)
and the false positive rate (specificity) at different thresholds
{[}57-58{]}. The area under the ROC curve (AUC) measures the MOD-1D-CNN
and MOD-2D-CNN model's ability to differentiate between classes.

\begin{enumerate}
\def\labelenumi{\Alph{enumi}.}
\setcounter{enumi}{5}
\item
  \textbf{Confusion Matrix:}
\end{enumerate}

The confusion matrix is a tabular representation of the MOD-1D-CNN and
MOD-2D-CNN model's predicted classifications against the actual ground
truth, presenting the counts of true Demented cases, true Non-Demented
cases, falsely predicted Demented cases, and falsely predicted
Non-Demented cases, and serves as the basis for calculating accuracy,
precision, recall, and F1-score. The confusion matrix helps analyze the
MOD-1D-CNN and MOD-2D-CNN model's errors, identify areas for
improvement, and optimize its performance. It provides a comprehensive
understanding of the model's classification capabilities, enabling
further enhancements and optimizations {[}59-62{]}. This enhances the
accuracy and reliability of the model, benefiting patients and
healthcare providers.

\textbf{4.2 Result Analysis of MOD-1D-CNN}

The Mod 1D CNN model's classification report shows that it has a
remarkable capacity to distinguish between the Demented and Non-Demented
classes. The performance of the MOD-1D Convolutional Neural Network
(CNN) model is evaluated across different epochs to understand its
learning trajectory. After training for 100 epochs, the model achieved
an accuracy of 0.62, precision of 0.59, recall of 0.61, and an F1-score
of 0.61. As training progressed to 150 epochs, there was a noticeable
improvement in accuracy (0.64), precision (0.62), recall (0.61), and
F1-score (0.62). Finally, after 200 epochs, the model demonstrated
further enhancement, achieving an accuracy of 0.70, precision of 0.68,
recall of 0.69, and an F1-score of 0.68. The accuracy, which represents
the overall correctness of the model's predictions, exhibited a
consistent upward trend, reaching its highest value at 200 epochs.
Precision, indicating the model's ability to avoid false positives,
increased steadily. Recall, highlighting the model's capacity to capture
true positive instances, also showed improvement over the epochs. The
F1-score, a harmonic mean of precision and recall, reached 0.68,
emphasizing the balanced performance of the model in identifying
instances of dementia. A thorough analysis of the model's performance
across different epochs can be found in the classification report in
Table 7.
\begin{center}

\textbf{Table 7:} Performance metrics evolution of Mod 1D CNN Model
\end{center}
\begin{longtable}[]{@{}
  >{\raggedright\arraybackslash}p{(\columnwidth - 8\tabcolsep) * \real{0.1999}}
  >{\raggedright\arraybackslash}p{(\columnwidth - 8\tabcolsep) * \real{0.2000}}
  >{\raggedright\arraybackslash}p{(\columnwidth - 8\tabcolsep) * \real{0.2000}}
  >{\raggedright\arraybackslash}p{(\columnwidth - 8\tabcolsep) * \real{0.2000}}
  >{\raggedright\arraybackslash}p{(\columnwidth - 8\tabcolsep) * \real{0.2000}}@{}}
\toprule()
\begin{minipage}[b]{\linewidth}\raggedright
\textbf{Epoch}
\end{minipage} & \begin{minipage}[b]{\linewidth}\raggedright
\textbf{Accuracy}
\end{minipage} & \begin{minipage}[b]{\linewidth}\raggedright
\textbf{Precision}
\end{minipage} & \begin{minipage}[b]{\linewidth}\raggedright
\textbf{Recall}
\end{minipage} & \begin{minipage}[b]{\linewidth}\raggedright
\textbf{F1-Score}
\end{minipage} \\
\midrule()
\endhead
\textbf{100} & 0.62 & 0.59 & 0.61 & 0.61 \\
\textbf{150} & 0.64 & 0.62 & 0.61 & 0.62 \\
\textbf{200} & 0.70 & 0.68 & 0.69 & 0.68 \\
\bottomrule()
\end{longtable}

Fig. 9, shows how training accuracy and validation accuracy changed
against epochs for the MOD-1D-CNN model.
\begin{center}

\includegraphics[width=4.14144in,height=3.05215in]{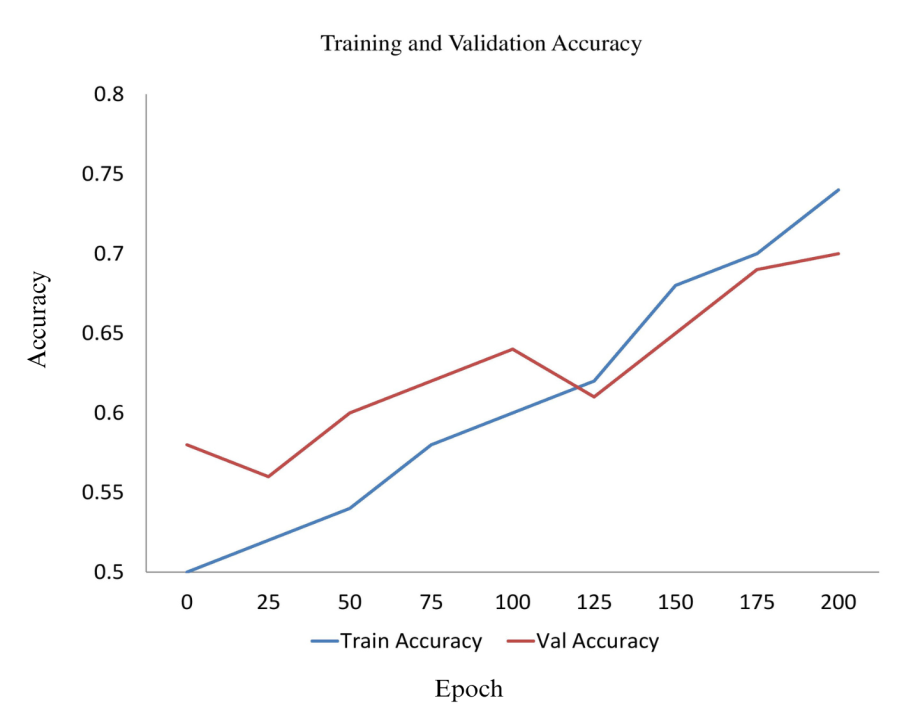}

\textbf{Fig. 9} Accuracy vs Epoch
\end{center}

Likewise, Fig. 10 shows how loss and validation loss changed against
epochs for the MOD-1D-CNN model.

\begin{center}

\includegraphics[width=4.36136in,height=3.28847in]{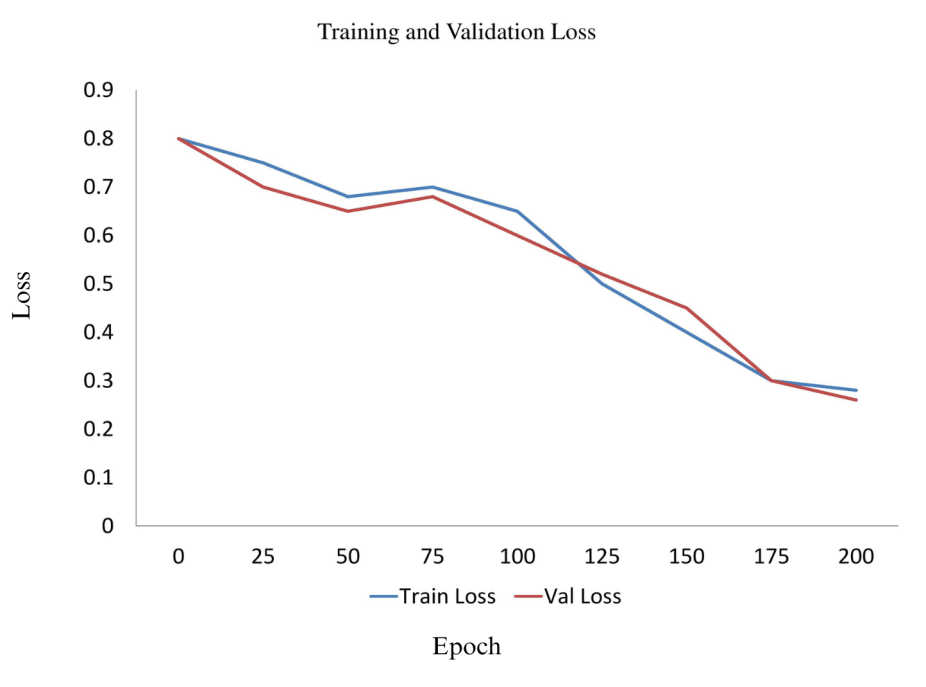}

\textbf{Fig. 10} Loss vs Epoch

\end{center}

\textbf{4.3 Result Analysis of MOD-2D-CNN}

In Table 7, we present the comprehensive results of the MOD-2D-CNN model
for facial image analysis, specifically focused on distinguishing
between demented and non-demented classes. The model's performance
metrics are evaluated across different training epochs to understand its
evolving effectiveness. In the first trial at epoch 30, the MOD-2D-CNN
model demonstrated a commendable performance with an accuracy of 86\%.
Precision, a metric emphasizing the model's capability to minimize false
positives, is observed at 86\%, indicating accurate positive
predictions. The recall, representing the model's effectiveness in
identifying all relevant instances of a class, is also at 86\%, ensuring
comprehensive coverage of both demented and non-demented instances. The
F1-score, a harmonic mean of precision and recall, stood at 86\%. In the
second trial at epoch 40, the model continued to show improvement. The
accuracy experienced a slight increase to 87\%, reinforcing the model's
capacity for precise classification. Precision remained high at 86\%,
underscoring the model's consistency in making accurate positive
predictions. However, there is a modest reduction in recall to 84\%. The
F1-score at this epoch is 84\%. A detailed analysis of the
classification report across different epochs is shown in Table 8.
\begin{center}

\textbf{Table 8:} Performance metrics of MOD-2D-CNN
\end{center}
\begin{longtable}[]{@{}
  >{\raggedright\arraybackslash}p{(\columnwidth - 8\tabcolsep) * \real{0.2021}}
  >{\raggedright\arraybackslash}p{(\columnwidth - 8\tabcolsep) * \real{0.1979}}
  >{\raggedright\arraybackslash}p{(\columnwidth - 8\tabcolsep) * \real{0.2000}}
  >{\raggedright\arraybackslash}p{(\columnwidth - 8\tabcolsep) * \real{0.2000}}
  >{\raggedright\arraybackslash}p{(\columnwidth - 8\tabcolsep) * \real{0.2000}}@{}}
\toprule()
\begin{minipage}[b]{\linewidth}\raggedright
\textbf{Epoch}
\end{minipage} & \begin{minipage}[b]{\linewidth}\raggedright
\textbf{Accuracy}
\end{minipage} & \begin{minipage}[b]{\linewidth}\raggedright
\textbf{Precision}
\end{minipage} & \begin{minipage}[b]{\linewidth}\raggedright
\textbf{Recall}
\end{minipage} & \begin{minipage}[b]{\linewidth}\raggedright
\textbf{F1-Score}
\end{minipage} \\
\midrule()
\endhead
\textbf{30} & 0.86 & 0.86 & 0.86 & 0.86 \\
\textbf{40} & 0.87 & 0.86 & 0.84 & 0.84 \\
\textbf{50} & 0.95 & 0.94 & 0.93 & 0.93 \\
\bottomrule()
\end{longtable}

In the final trial at epoch 50, the model reached its peak performance,
demonstrating an impressive accuracy of 95\%. This substantial
improvement underscored the model's advanced ability to correctly
categorize instances. Precision maintained a high level of 94\%,
indicating the model's continued accuracy in positive predictions.
Notably, recall experienced a substantial improvement to 93\%,
signifying the model's heightened capability to identify instances
across both classes. The F1-score at this epoch reached 93\%. Looking at
the confusion matrix in Fig. 11, we can see that there are 164 truly
predicted demented cases (TP), where the model correctly classified
demented cases. There are 171 truly predicted non-demented cases (TN),
where the model correctly classified non-demented cases. The model has 9
falsely predicted demented cases (FP), incorrectly classifying
non-demented cases as demented. Additionally, there are 16 falsely
predicted non-demented cases (FN), where the model incorrectly
classified demented cases as non-demented.

\begin{center}

\includegraphics[width=3.21337in,height=2.26158in]{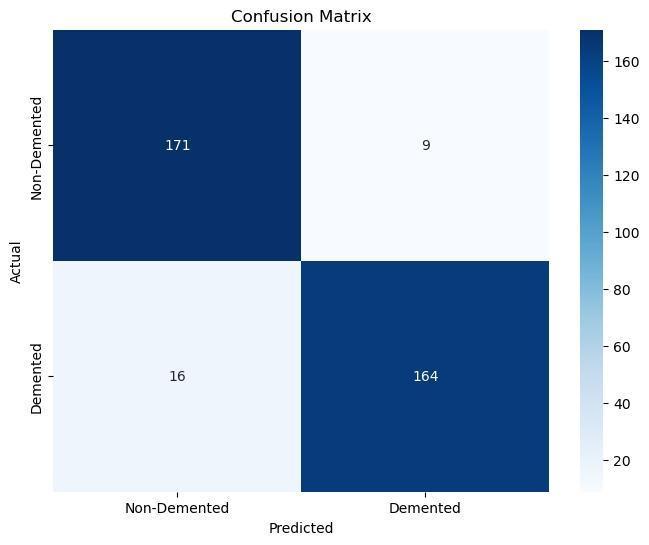}

\textbf{Fig. 11} Confusion Matrix of the MOD-2D-CNN Model

\end{center}

The accuracy versus epoch plot, presented below in Fig. 12, provides a
dynamic visual representation of the MOD-2D-CNN model's performance
throughout 50 epochs during the training process.

\begin{center}

\includegraphics[width=3.18238in,height=2.31049in]{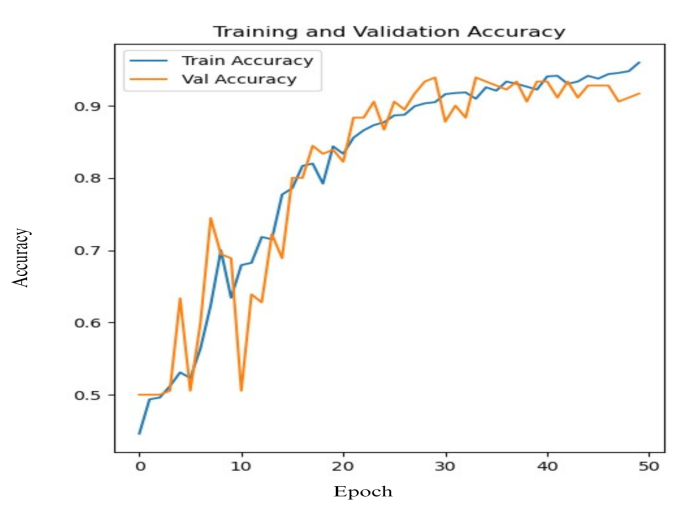}

\textbf{Fig. 12} Accuracy vs Epochs
\end{center}

Similarly, Fig. 13 shows training loss and validation loss of the
MOD-2D-CNN models across 50 epochs.

\begin{center}

\includegraphics[width=3.29219in,height=2.04955in]{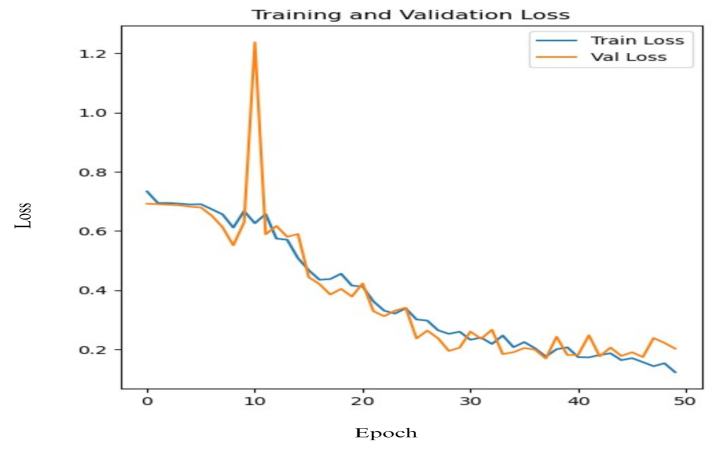}

\textbf{Fig. 13} Loss vs Epochs
\end{center}

An Area under the ROC Curve (AUC) of 0.97 in Fig. 14 shown below
provides a visual representation of the MOD-2D-CNN model's performance
which indicates that the Receiver Operating Characteristic (ROC) curve
performs well in distinguishing between demented and non-demented
individuals based on facial images. This represents that the model is
quite accurate in differentiating between those who have dementia and
those who do not.

\begin{center}

\includegraphics[width=3.22307in,height=1.86472in]{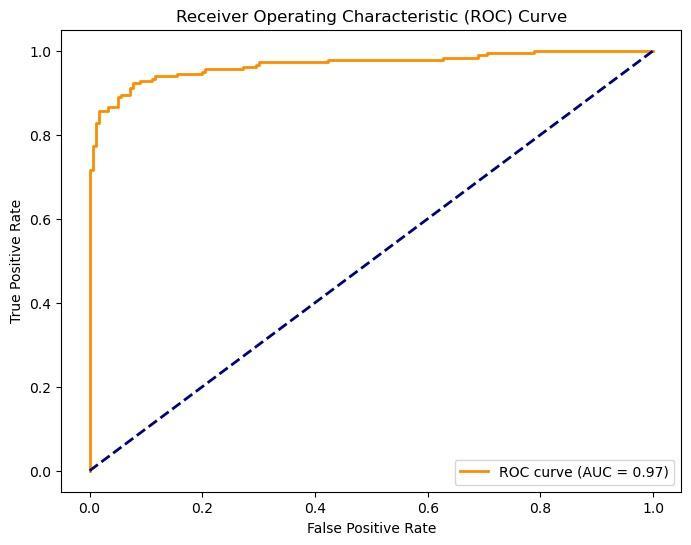}

\textbf{Fig. 14} Receiver Operating Characteristic (ROC) Curve for
MOD-2D-CNN Model

\end{center}

To guarantee the predictive capacity for the face analysis of demented
and non-demented individuals, a thorough assessment of the MOD-2D-CNN
model's predictive performance is carried out utilizing a collection of
test photos. The MOD-2D-CNN model demonstrated its proficiency in
analyzing facial features to make predictions about cognitive health.
The resulting predictions of the test images are shown in Fig. 15. This
analysis shows the model's effectiveness in discerning subtle patterns
within facial imagery, underlining its potential as a valuable tool for
early diagnosis and risk assessment in the context of dementia. The
results presented herein exemplify a dual-model strategy for dementia
prediction, with MOD-1D-CNN holding a 30\% weightage and MOD-2D-CNN
taking precedence with a substantial 70\% weightage.

\begin{center}

\includegraphics[width=4.46605in,height=2.40866in]{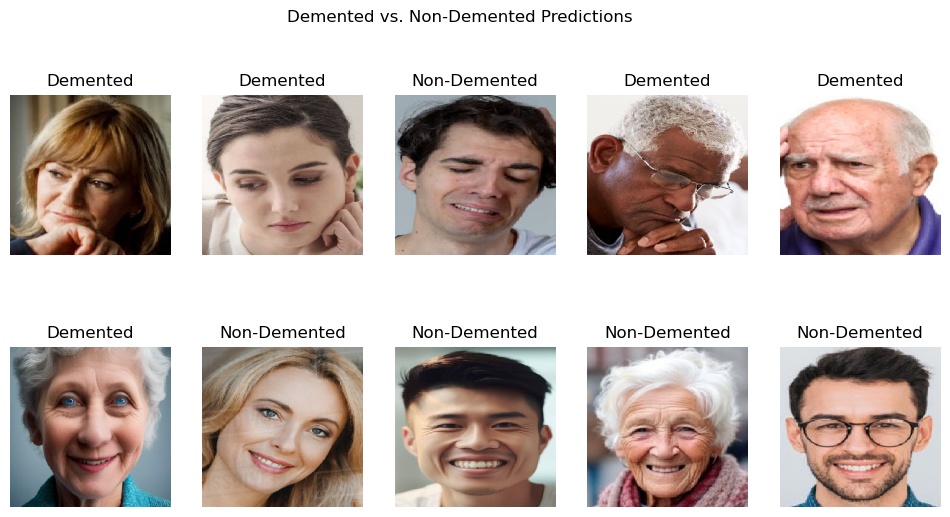}

\textbf{Fig. 15} Predictive Analysis of Dementia Using Facial Images

\end{center}

In Fig. 16, MOD-1D-CNN predicts a case as Demented using typical test
health metrics data, and MOD-2D-CNN aligns with this prediction using a
facial image, resulting in an overall decision of Demented. In Fig. 17,
although MOD-1D-CNN predicts Non-Demented, the ultimate decision is in
favor of demented due to the higher confidence level provided by
MOD-2D-CNN.

\textbf{4.4 Discussion and Comparison}

\textbf{4.3.1 MOD-1D-CNN}

In the comparison analysis of MOD-1D-CNN, various optimizers, namely
Adam, Adadelta, Adamax, Adagrad, SGD, RMSprop, and AdamW, are employed
to evaluate the model's performance in terms of accuracy, precision,
recall, and F1-score, as demonstrated in Table 9. The findings reveal
that the Adam optimizer achieved the highest accuracy of 0.70,
surpassing other optimizers. In terms of precision, Adam also
outperformed other optimizers with a precision score of 0.68, while
other optimizers exhibited the lowest precision.

\begin{center}

\includegraphics[width=4.18056in,height=3.30208in]{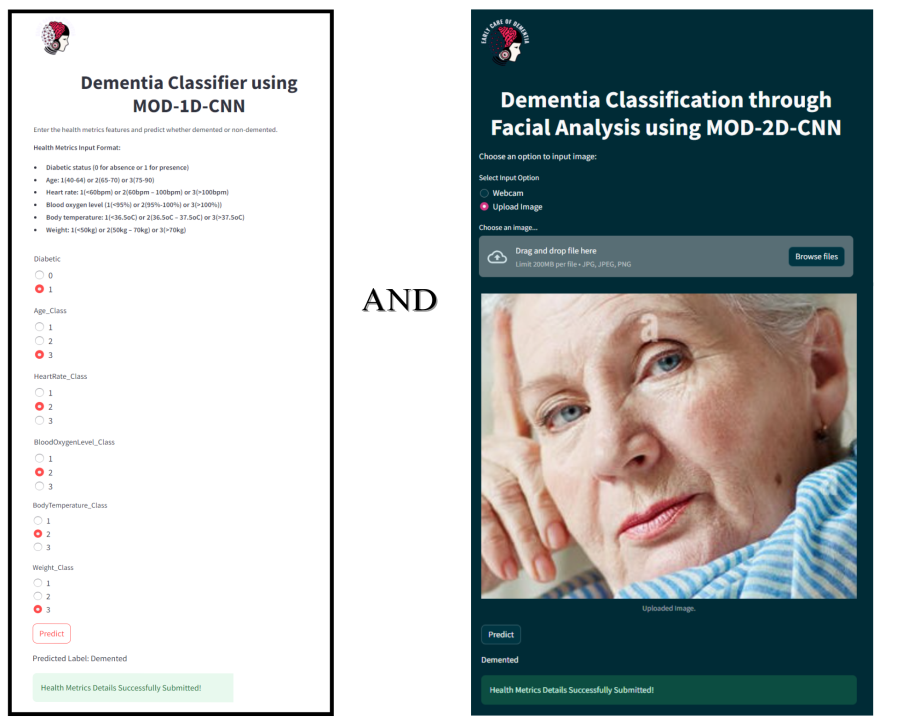}\includegraphics[width=0.32153in,height=0.38194in]{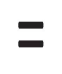}\includegraphics[width=1.49514in,height=0.8in]{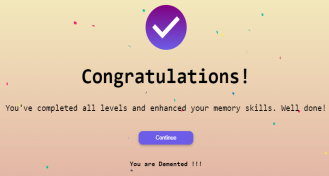}

\textbf{Fig. 16} Decision Outcome - Demented with Congratulations Page
\end{center}

A similar trend is observed in the recall, where Adam demonstrated the
highest score of 0.69, while all others are at the lowest. F1-score,
reflecting the balance between precision and recall, mirrored the
overall trends, with Adam leading with 0.68. This suggests that Adam's
optimization is better for understanding the patterns in the dataset for
the accurate prediction of the class labels demented or non-demented.
\begin{center}

\textbf{Table 9:} Comparison of Training MOD-1D-CNN using Different
Optimizers
\end{center}
\begin{longtable}[]{@{}
  >{\raggedright\arraybackslash}p{(\columnwidth - 14\tabcolsep) * \real{0.1212}}
  >{\raggedright\arraybackslash}p{(\columnwidth - 14\tabcolsep) * \real{0.0971}}
  >{\raggedright\arraybackslash}p{(\columnwidth - 14\tabcolsep) * \real{0.1431}}
  >{\raggedright\arraybackslash}p{(\columnwidth - 14\tabcolsep) * \real{0.1229}}
  >{\raggedright\arraybackslash}p{(\columnwidth - 14\tabcolsep) * \real{0.1532}}
  >{\raggedright\arraybackslash}p{(\columnwidth - 14\tabcolsep) * \real{0.1061}}
  >{\raggedright\arraybackslash}p{(\columnwidth - 14\tabcolsep) * \real{0.1448}}
  >{\raggedright\arraybackslash}p{(\columnwidth - 14\tabcolsep) * \real{0.1117}}@{}}
\toprule()
\begin{minipage}[b]{\linewidth}\raggedright
\end{minipage} & \begin{minipage}[b]{\linewidth}\raggedright
\end{minipage} & \begin{minipage}[b]{\linewidth}\raggedright
\end{minipage} & \begin{minipage}[b]{\linewidth}\raggedright
\end{minipage} & \begin{minipage}[b]{\linewidth}\raggedright
\textbf{Optimizers}
\end{minipage} & \begin{minipage}[b]{\linewidth}\raggedright
\end{minipage} & \begin{minipage}[b]{\linewidth}\raggedright
\end{minipage} & \begin{minipage}[b]{\linewidth}\raggedright
\end{minipage} \\
\midrule()
\endhead
& \textbf{Adam} & \textbf{Adadelta} & \textbf{Adamax} & \textbf{Adagrad}
& \textbf{SGD} & \textbf{RMSprop} & \textbf{AdamW} \\
\textbf{Accuracy} & 0.70 & 0.50 & 0.62 & 0.49 & 0.55 & 0.62 & 0.55 \\
\textbf{Precision} & 0.68 & 0.48 & 0.58 & 0.47 & 0.53 & 0.59 & 0.52 \\
\textbf{Recall} & 0.69 & 0.49 & 0.60 & 0.47 & 0.52 & 0.60 & 0.54 \\
\textbf{F1-Score} & 0.68 & 0.47 & 0.59 & 0.47 & 0.52 & 0.60 & 0.54 \\
\bottomrule()
\end{longtable}

\begin{center}

\includegraphics[width=4.17292in,height=3.64514in]{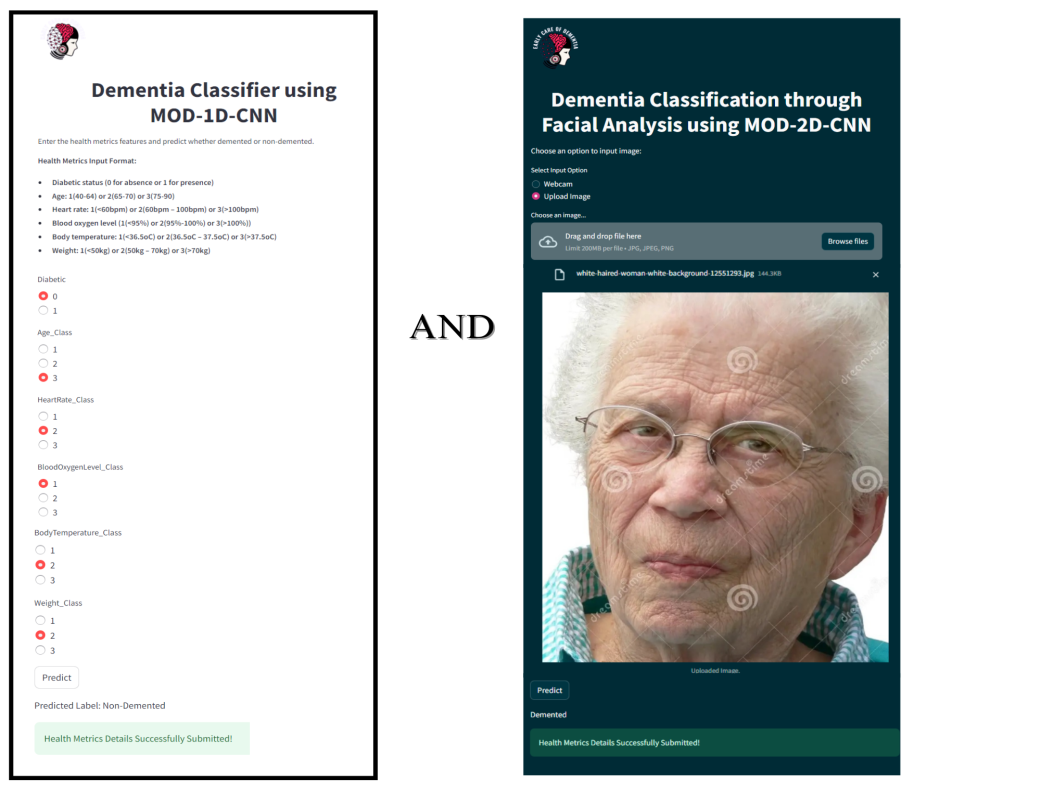}\includegraphics[width=0.34722in,height=0.35625in]{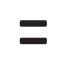}\includegraphics[width=1.7125in,height=1in]{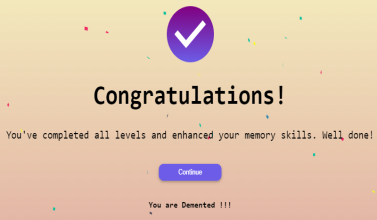}

\textbf{Fig. 17} Decision Outcome - Demented with high probability
\end{center}

In Fig. 18, although MOD-1D-CNN predicts Demented, the ultimate decision
is in favor of Non-Demented due to the higher confidence level provided
by MOD-2D-CNN.

\begin{center}

\includegraphics[width=4.41181in,height=3.31736in]{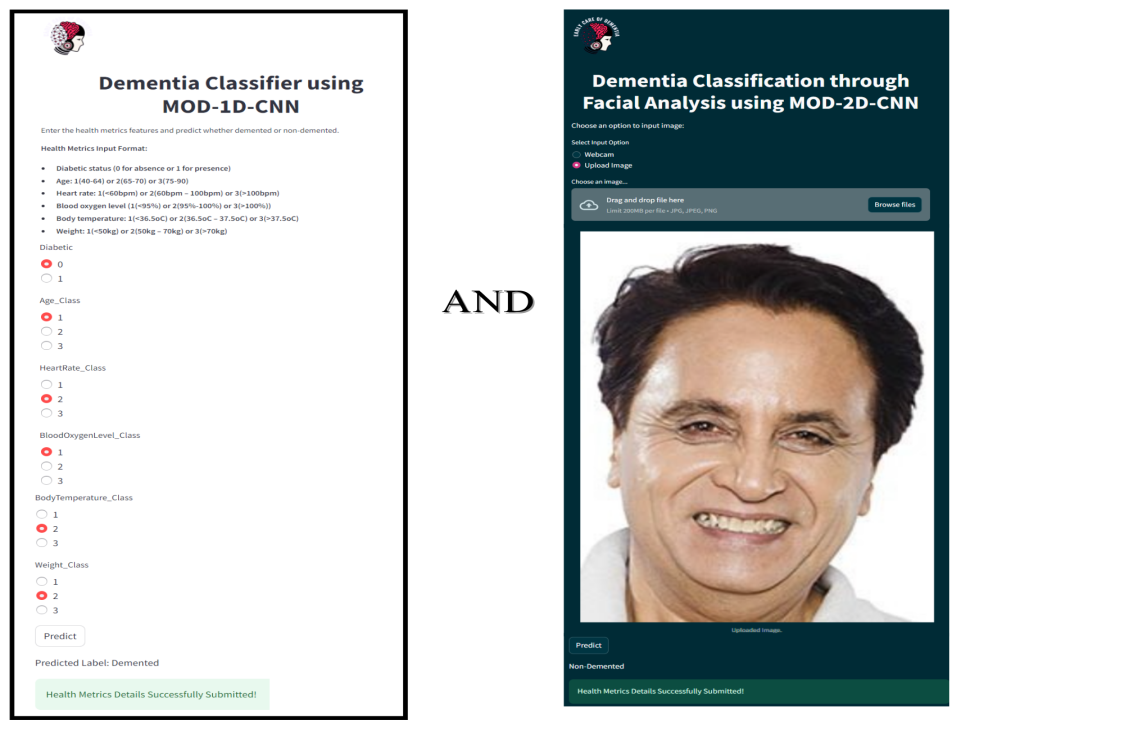}\includegraphics[width=0.39063in,height=0.39063in]{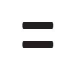}\includegraphics[width=1.40833in,height=0.84306in]{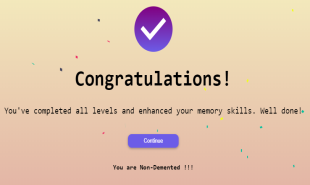}

\textbf{Fig. 18} Decision Outcome - Non-Demented with high probability

\end{center}

Conversely, Fig. 19 showcases agreement between both models in
predicting Non-Demented, resulting in a conclusive decision of
Non-Demented.

\begin{center}

\includegraphics[width=4.20069in,height=3.80625in]{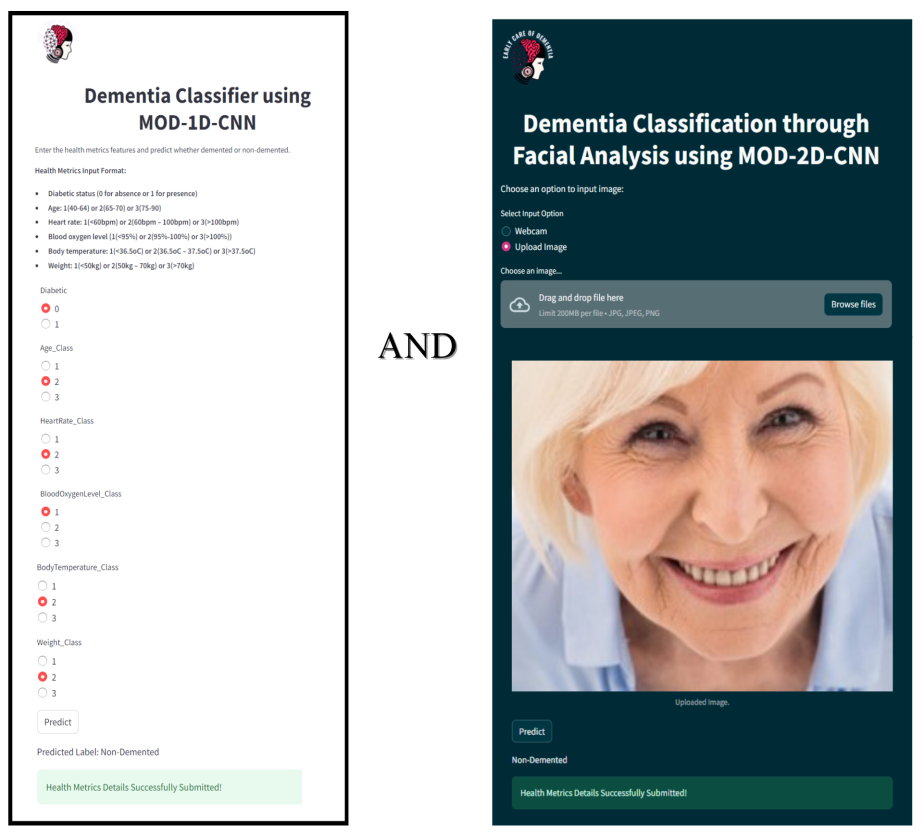}\includegraphics[width=0.39063in,height=0.39063in]{vertopal_2526c9e576254fc38e49376f43a98ce3/media/image27.png}\includegraphics[width=1.55625in,height=0.81736in]{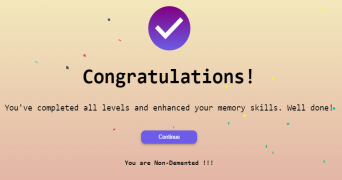}

\textbf{Fig. 19} Decision Outcome - Non-Demented
\end{center}

The findings indicate that the MOD-1D-CNN's performance is greatly
impacted by the optimizer selected, with Adam showing to be the most
efficient in terms of all evaluation metrics. This investigation sheds
light on the specifics of optimizer influence on model performance and
emphasizes how crucial it is to use the right optimizer for 1D
convolutional neural networks to get optimal results. In the comparative
analysis of model parameters for MOD-1D-CNN, artificial neural network
(ANN), and long short-term memory (LSTM), the focus lies on trainable
and non-trainable parameters, as well as the overall total parameters. A
comparison of parameters to those of other various models is shown in
Table 10. MOD-1D-CNN exhibits a total of 21,550 parameters, with all
being trainable, signifying the model's capability to learn from the
training data. In contrast, the artificial neural network (ANN)
demonstrates a significantly lower parameter count, with a total of 163
trainable parameters. In this case, there are no non-trainable
parameters, emphasizing the simplicity of the model. Long short-term
memory (LSTM), on the other hand, presents a larger parameter count,
with 120,641 trainable parameters and no non-trainable parameters.
\begin{center}

\textbf{Table 10:} Comparison of the Number of Parameters of the
MOD-1D-CNN model with other models
\end{center}
\begin{longtable}[]{@{}
  >{\raggedright\arraybackslash}p{(\columnwidth - 6\tabcolsep) * \real{0.2111}}
  >{\raggedright\arraybackslash}p{(\columnwidth - 6\tabcolsep) * \real{0.2667}}
  >{\raggedright\arraybackslash}p{(\columnwidth - 6\tabcolsep) * \real{0.2889}}
  >{\raggedright\arraybackslash}p{(\columnwidth - 6\tabcolsep) * \real{0.2333}}@{}}
\toprule()
\begin{minipage}[b]{\linewidth}\raggedright
\textbf{Models}
\end{minipage} & \begin{minipage}[b]{\linewidth}\raggedright
\textbf{Trainable Parameters}
\end{minipage} & \begin{minipage}[b]{\linewidth}\raggedright
\textbf{Non-Trainable Parameters}
\end{minipage} & \begin{minipage}[b]{\linewidth}\raggedright
\textbf{Total Parameters}
\end{minipage} \\
\midrule()
\endhead
MOD-1D-CNN & 21550 & 0 & 21,550 \\
ANN & 163 & 0 & 163 \\
LSTM & 120641 & 0 & 1,20,641 \\
\bottomrule()
\end{longtable}

This analysis underscores the diversity in parameter complexity across
the models. MOD-1D-CNN's intermediate parameter count suggests a balance
between model complexity and simplicity. The effectiveness of the
MOD-1D-CNN model is compared with the other cutting-edge deep learning
models that are shown in Table 11. The training, validation, and test
accuracies are assessed at 100, 150, and 200 epochs in the thorough
comparison analysis across various training epochs for MOD-1D-CNN,
artificial neural network (ANN), and long short-term memory (LSTM) to
determine the model's learning capacities over time. At 100 epochs,
MOD-1D-CNN exhibits a training accuracy of 63.1\%, a validation accuracy
of 61.0\%, and a test accuracy of 58.9\%. In comparison, ANN achieves
slightly lower accuracies with 60.8\%, 58.5\%, and 59.2\% for training,
validation, and test sets, respectively. Long short-term memory (LSTM),
on the other hand, records accuracies of 59.8\%, 56.8\%, and 59.8\%. As
the training progresses to 150 epochs, MOD-1D-CNN sees improvements,
achieving training, validation, and test accuracies of 64.6\%, 62.6\%,
and 62.3\%, respectively. ANN also demonstrates incremental improvements
with accuracies of 61.7\%, 59.8\%, and 60.1\%, while LSTM maintains
relatively stable performance with 60.7\%, 58.9\%, and 60.0\% for the
three sets. Upon reaching 200 epochs, MOD-1D-CNN continues to exhibit
substantial enhancements, achieving training, validation, and test
accuracies of 73.2\%, 71.8\%, and 70.5\%. ANN displays a modest increase
with accuracies of 62.6\%, 64.5\%, and 61.2\%, while LSTM shows marginal
improvements, recording accuracies of 61.6\%, 61.3\%, and 60.5\%.
\begin{center}

\textbf{Table 11:} Comparison of Accuracy of MOD-1D-CNN with other
models for different epochs
\end{center}
\begin{longtable}[]{@{}
  >{\raggedright\arraybackslash}p{(\columnwidth - 18\tabcolsep) * \real{0.0910}}
  >{\raggedright\arraybackslash}p{(\columnwidth - 18\tabcolsep) * \real{0.0998}}
  >{\raggedright\arraybackslash}p{(\columnwidth - 18\tabcolsep) * \real{0.1140}}
  >{\raggedright\arraybackslash}p{(\columnwidth - 18\tabcolsep) * \real{0.0993}}
  >{\raggedright\arraybackslash}p{(\columnwidth - 18\tabcolsep) * \real{0.0993}}
  >{\raggedright\arraybackslash}p{(\columnwidth - 18\tabcolsep) * \real{0.1048}}
  >{\raggedright\arraybackslash}p{(\columnwidth - 18\tabcolsep) * \real{0.0952}}
  >{\raggedright\arraybackslash}p{(\columnwidth - 18\tabcolsep) * \real{0.0966}}
  >{\raggedright\arraybackslash}p{(\columnwidth - 18\tabcolsep) * \real{0.1021}}
  >{\raggedright\arraybackslash}p{(\columnwidth - 18\tabcolsep) * \real{0.0979}}@{}}
\toprule()
\begin{minipage}[b]{\linewidth}\raggedright
\end{minipage} & \begin{minipage}[b]{\linewidth}\raggedright
\end{minipage} & \begin{minipage}[b]{\linewidth}\raggedright
\textbf{100 Epochs}
\end{minipage} & \begin{minipage}[b]{\linewidth}\raggedright
\end{minipage} & \begin{minipage}[b]{\linewidth}\raggedright
\end{minipage} & \begin{minipage}[b]{\linewidth}\raggedright
\textbf{150 Epochs}
\end{minipage} & \begin{minipage}[b]{\linewidth}\raggedright
\end{minipage} & \begin{minipage}[b]{\linewidth}\raggedright
\end{minipage} & \begin{minipage}[b]{\linewidth}\raggedright
\textbf{200 Epochs}
\end{minipage} & \begin{minipage}[b]{\linewidth}\raggedright
\end{minipage} \\
\midrule()
\endhead
\textbf{Models}  & \textbf{Train Accuracy} & \textbf{Val Accuracy}
& \textbf{Test Accuracy} & \textbf{Train Accuracy} &
\textbf{Val Accuracy} & \textbf{Test Accuracy} & \textbf{Train
Accuracy} & \textbf{Val Accuracy} & \textbf{Test Accuracy} \\
MOD-1D-CNN & 0.631 & 0.610 & 0.589 & 0.646 & 0.626 & 0.623 & 0.732 &
0.718 & 0.705 \\
ANN & 0.608 & 0.585 & 0.592 & 0.617 & 0.598 & 0.601 & 0.626 & 0.645 &
0.612 \\
LSTM & 0.598 & 0.568 & 0.598 & 0.607 & 0.589 & 0.600 & 0.616 & 0.613 &
0.605 \\
\bottomrule()
\end{longtable}

This analysis provides insights into the model's learning trajectories
over different epochs, demonstrating MOD-1D-CNN's capacity for continued
improvement and outperforming both ANN and LSTM across various training
durations.

\textbf{4.3.2 MOD-2D-CNN}

The MOD-2D-CNN model is compiled using seven different optimizers for
comparison analysis across diverse optimizers to test on the image set,
and their performance is evaluated on a test set of 360 images.
\begin{center}

\textbf{Table 12:} Comparison of Training MOD-2D-CNN using Different
Optimizers
\end{center}
\begin{longtable}[]{@{}
  >{\raggedright\arraybackslash}p{(\columnwidth - 14\tabcolsep) * \real{0.1212}}
  >{\raggedright\arraybackslash}p{(\columnwidth - 14\tabcolsep) * \real{0.0971}}
  >{\raggedright\arraybackslash}p{(\columnwidth - 14\tabcolsep) * \real{0.1431}}
  >{\raggedright\arraybackslash}p{(\columnwidth - 14\tabcolsep) * \real{0.1229}}
  >{\raggedright\arraybackslash}p{(\columnwidth - 14\tabcolsep) * \real{0.1532}}
  >{\raggedright\arraybackslash}p{(\columnwidth - 14\tabcolsep) * \real{0.1061}}
  >{\raggedright\arraybackslash}p{(\columnwidth - 14\tabcolsep) * \real{0.1448}}
  >{\raggedright\arraybackslash}p{(\columnwidth - 14\tabcolsep) * \real{0.1117}}@{}}
\toprule()
\begin{minipage}[b]{\linewidth}\raggedright
\end{minipage} & \begin{minipage}[b]{\linewidth}\raggedright
\end{minipage} & \begin{minipage}[b]{\linewidth}\raggedright
\end{minipage} & \begin{minipage}[b]{\linewidth}\raggedright
\end{minipage} & \begin{minipage}[b]{\linewidth}\raggedright
\textbf{Optimizers}
\end{minipage} & \begin{minipage}[b]{\linewidth}\raggedright
\end{minipage} & \begin{minipage}[b]{\linewidth}\raggedright
\end{minipage} & \begin{minipage}[b]{\linewidth}\raggedright
\end{minipage} \\
\midrule()
\endhead
& \textbf{Adam} & \textbf{Adadelta} & \textbf{Adamax} & \textbf{Adagrad}
& \textbf{SGD} & \textbf{RMSprop} & \textbf{AdamW} \\
\textbf{Accuracy} & 0.95 & 0.55 & 0.83 & 0.51 & 0.50 & 0.92 & 0.88 \\
\textbf{Precision} & 0.93 & 0.54 & 0.83 & 0.50 & 0.49 & 0.92 & 0.89 \\
\textbf{Recall} & 0.93 & 0.50 & 0.83 & 0.50 & 0.48 & 0.92 & 0.88 \\
\textbf{F1-Score} & 0.93 & 0.50 & 0.83 & 0.50 & 0.48 & 0.92 & 0.87 \\
\bottomrule()
\end{longtable}

The optimizer selection has a considerable impact on the MOD-2D-CNN
model's performance, as demonstrated in Table 12. Among the optimizers
tested during 50 epochs of iteration, Adam emerged as the most
effective, achieving the highest accuracy of 0.95 and demonstrating
strong precision, recall, and f1-score all hovering around values of
0.93 at 50 epochs. This suggests that Adam's optimization is well-suited
for enhancing the model's ability to correctly classify both positive
and negative instances, as well as minimizing false positives and false
negatives. Adam's adaptability makes it an effective choice for deep
learning tasks such as image classification since it allows it to
converge rapidly and efficiently over complicated loss landscapes. Table
13 compares the MOD-2D-CNN model's total number of parameters to those
of various well-known pre-trained models, including VGG-16, ResNet-50,
and MobileNet. The MOD-2D-CNN model demonstrates an impressive parameter
efficiency with a total of 19,394,881 parameters consisting entirely of
trainable parameters, indicating that its architecture does not
incorporate non-trainable parameters and the value for non-trainable
parameters is 0. In contrast, VGG16, a well-known architecture, has
21,137,729 total parameters which is little much more than the
MOD-2D-CNN model, with 6,423,041 being trainable and 14,714,688
non-trainable. ResNet50 surpasses the others in terms of the sheer
number of parameters, with 49,278,337 in total, comprising 25,690,625
trainable parameters and 23,587,712 non-trainable parameters which is
two and a half times more than the MOD-2D-CNN model. MobileNet, on the
other hand, has a total of 16,074,433 parameters in total, which is
somehow less than the MOD-2D-CNN model's total parameters because of its
lightweight architecture which is reflected in its lower parameter count
making it suitable for resource-constrained architecture, with
12,845,569 being trainable and 3,228,864 non-trainable.
\begin{center}

\textbf{Table 13:} Comparison of the Number of Parameters of the
MOD-2D-CNN model with other models
\end{center}
\begin{longtable}[]{@{}
  >{\raggedright\arraybackslash}p{(\columnwidth - 6\tabcolsep) * \real{0.2501}}
  >{\raggedright\arraybackslash}p{(\columnwidth - 6\tabcolsep) * \real{0.2500}}
  >{\raggedright\arraybackslash}p{(\columnwidth - 6\tabcolsep) * \real{0.2500}}
  >{\raggedright\arraybackslash}p{(\columnwidth - 6\tabcolsep) * \real{0.2500}}@{}}
\toprule()
\begin{minipage}[b]{\linewidth}\raggedright
\textbf{Models}
\end{minipage} & \begin{minipage}[b]{\linewidth}\raggedright
\textbf{Trainable Parameters}
\end{minipage} & \begin{minipage}[b]{\linewidth}\raggedright
\textbf{Non-Trainable}

\textbf{Parameters}
\end{minipage} & \begin{minipage}[b]{\linewidth}\raggedright
\textbf{Total}

\textbf{Parameters}
\end{minipage} \\
\midrule()
\endhead
MOD-2D-CNN & 19,394,881 & 0 & 19,394,881 \\
VGG16 & 6,423,041 & 14,714,688 & 21,137,729 \\
RestNet50 & 25,690,625 & 23,587,712 & 49,278,337 \\
MobileNet & 12,845,569 & 3,228,864 & 16,074,433 \\
\bottomrule()
\end{longtable}

In contrast, the MOD-2D-CNN model's design strikes a balance between
model complexity and parameter efficiency, showcasing that high accuracy
and performance can be achieved without excessive parameter usage.
MOD-2D-CNN parameter efficiency also extends its attention to the
non-trainable parameters because its value is 0 which says that
architecture does not exhibit any non-trainable parameters. In contrast,
other pre-trained models like ResNet50 have a staggering 23 million
non-trainable parameters, which can lead to increased memory usage and
longer initialization times. This simplicity makes MOD-2D-CNN a more
practical choice for resource-constrained environments, such as edge
devices or low-power hardware. Moreover, a model with fewer parameters
tends to generalize better and is less prone to overfitting. MOD-2D-CNN
relatively smaller parameter count allows it to generalize well even
with limited training data, making it a robust and reliable solution for
dementia classification. While some of the larger pre-trained models
like VGG16, ResNet50, and MobileNet might achieve slightly higher
accuracy on certain datasets, they come at the cost of increased
computational requirements and memory consumption. The parameter
efficiency of MOD-2D-CNN, combined with its competitive accuracy and
performance, positions it as a more noble and practical choice for
real-world applications. The effectiveness of the MOD-2D-CNN model is
also contrasted with that of other cutting-edge deep learning models,
such as the ones in Table 14 (VGG-16, ResNet50, and MobileNet). All the
models are trained on the same facial image dataset, allowing for a fair
and meaningful comparison across different epoch trials (30, 40, and
50).
\begin{center}

\textbf{Table 14:} Comparison of Accuracy of MOD-1D-CNN with other
pre-trained models with different epochs
\end{center}
\begin{longtable}[]{@{}
  >{\raggedright\arraybackslash}p{(\columnwidth - 18\tabcolsep) * \real{0.1010}}
  >{\raggedright\arraybackslash}p{(\columnwidth - 18\tabcolsep) * \real{0.1051}}
  >{\raggedright\arraybackslash}p{(\columnwidth - 18\tabcolsep) * \real{0.1079}}
  >{\raggedright\arraybackslash}p{(\columnwidth - 18\tabcolsep) * \real{0.0982}}
  >{\raggedright\arraybackslash}p{(\columnwidth - 18\tabcolsep) * \real{0.0968}}
  >{\raggedright\arraybackslash}p{(\columnwidth - 18\tabcolsep) * \real{0.1010}}
  >{\raggedright\arraybackslash}p{(\columnwidth - 18\tabcolsep) * \real{0.0968}}
  >{\raggedright\arraybackslash}p{(\columnwidth - 18\tabcolsep) * \real{0.0954}}
  >{\raggedright\arraybackslash}p{(\columnwidth - 18\tabcolsep) * \real{0.1010}}
  >{\raggedright\arraybackslash}p{(\columnwidth - 18\tabcolsep) * \real{0.0968}}@{}}
\toprule()
\begin{minipage}[b]{\linewidth}\raggedright
\end{minipage} & \begin{minipage}[b]{\linewidth}\raggedright
\end{minipage} & \begin{minipage}[b]{\linewidth}\raggedright
\textbf{30 Epochs}
\end{minipage} & \begin{minipage}[b]{\linewidth}\raggedright
\end{minipage} & \begin{minipage}[b]{\linewidth}\raggedright
\end{minipage} & \begin{minipage}[b]{\linewidth}\raggedright
\textbf{40 Epochs}
\end{minipage} & \begin{minipage}[b]{\linewidth}\raggedright
\end{minipage} & \begin{minipage}[b]{\linewidth}\raggedright
\end{minipage} & \begin{minipage}[b]{\linewidth}\raggedright
\textbf{50 Epochs}
\end{minipage} & \begin{minipage}[b]{\linewidth}\raggedright
\end{minipage} \\
\midrule()
\endhead
\textbf{Models} & \textbf{Train Accuracy} & \textbf{Val
Accuracy} & \textbf{Test Accuracy} & \textbf{Train Accuracy} &
\textbf{Val Accuracy} & \textbf{Test Accuracy} & \textbf{Train
Accuracy} & \textbf{Val Accuracy} & \textbf{Test Accuracy} \\
MOD-2D-CNN & 0.837 & 0.821 & 0.858 & 0.862 & 0.855 & 0.850 & 0.959 &
0.933 & 0.957 \\
VGG16 & 0.851 & 0.916 & 0.927 & 0.856 & 0.922 & 0.930 & 0.944 & 0.938 &
0.913 \\
Res\\Net50 & 0.50 & 0.50 & 0.50 & 0.51 & 0.52 & 0.50 & 0.50 & 0.50 &
0.50 \\
Mobile\\Net & 0.967 & 0.950 & 0.933 & 0.850 & 0.933 & 0.933 & 0.50 & 0.51
& 0.50 \\
\bottomrule()
\end{longtable}

In the case of MOD-2D-CNN, it demonstrated consistent improvement in
both training and validation accuracy as the number of epochs increased.
At 30 epochs, MOD-2D-CNN achieved a training accuracy of 83.7\%, a
validation accuracy of 82.1\%, and a test accuracy of 85.8\%. The model
continued to exhibit enhanced performance at 40 epochs, with training
accuracy reaching 86.2\%, validation accuracy at 85.5\%, and a test
accuracy of 85.0\%. Surprisingly, at 50 epochs, MOD-2D-CNN demonstrated
remarkable improvement, reaching a training accuracy of 95.9\%,
validation accuracy of 93.3\%, and test accuracy of 95.7\%. This notable
performance may indicate that the model benefits from increased training
epochs. Comparatively, VGG16 exhibited commendable performance across
different training epochs. At 30 epochs, the model achieved a training
accuracy of 85.1\%, a validation accuracy of 91.6\%, and a test accuracy
of 92.7\%. The accuracy levels continued to be consistent at 40 epochs,
with training accuracy reaching 85.6\%, validation accuracy at 92.2\%,
and test accuracy at 93.0\%. While there is a slight dip in test
accuracy at 50 epochs at 91.3\%, VGG16 maintained high training accuracy
at 94.4\% and validation accuracy at 93.8\%. This consistent performance
suggests that VGG16 is a robust model for image classification tasks.
However, MOD-2D-CNN demonstrated slightly superior accuracy at the same
epoch, suggesting its potential as a robust model for image
classification tasks. On the other hand, ResNet50 faced challenges in
achieving meaningful accuracy, struggling to surpass 50\% throughout all
epochs. At 30, 40, and 50 epochs, the model consistently reported a
training, validation, and test accuracy of 50.0\%, indicating potential
issues in convergence or model suitability for the dataset. This shows
that MOD-2D-CNN is more robust than ResNet50 in learning the patterns of
images for the accurate classification of demented people as well as
non-demented ones. Whereas MobileNet, while initially displaying
impressive performance at 30 epochs with a training accuracy of 96.7\%
and validation accuracy of 95.0\%, struggled to maintain accuracy in
subsequent epochs. At 40 epochs, the model's training accuracy dropped
to 85.0\%, and the validation accuracy decreased to 93.3\%, with the
test accuracy falling to 50.0\%. This trend continued at 50 epochs, with
training accuracy at 50.0\%, validation accuracy at 51.0\%, and test
accuracy at 50.0\%. In contrast, MOD-2D-CNN consistently outperformed
MobileNet across all epochs, highlighting its robustness in comparison.
In summary, MOD-2D-CNN demonstrated competitive accuracy levels and
exhibited excellent generalization, parameter efficiency, and robustness
in classifying demented as well as non-demented individuals through
facial analysis, especially at 50 epochs, making it a promising
candidate for image classification tasks when compared to the
competitive models like VGG16, ResNet50, and MobileNet. MOD-2D-CNN
consistently outperforms them on the test set which makes it a noble
choice and a leading solution for the classification of dementia through
facial image, highlighting its practicality and effectiveness in
real-world scenarios.
\begin{center}

\textbf{Table 15:} Comparison of the proposed methodology with some
popular gaming applications for healthcare

\end{center}

\begin{longtable}[]{@{}
  >{\raggedright\arraybackslash}p{(\columnwidth - 8\tabcolsep) * \real{0.2000}}
  >{\raggedright\arraybackslash}p{(\columnwidth - 8\tabcolsep) * \real{0.2000}}
  >{\raggedright\arraybackslash}p{(\columnwidth - 8\tabcolsep) * \real{0.2604}}
  >{\raggedright\arraybackslash}p{(\columnwidth - 8\tabcolsep) * \real{0.1562}}
  >{\raggedright\arraybackslash}p{(\columnwidth - 8\tabcolsep) * \real{0.1833}}@{}}
\toprule()
\begin{minipage}[b]{\linewidth}\raggedright
\textbf{Literature Works}
\end{minipage} & \begin{minipage}[b]{\linewidth}\raggedright
\textbf{Used Technology}
\end{minipage} & \begin{minipage}[b]{\linewidth}\raggedright
\textbf{Applications}
\end{minipage} & \begin{minipage}[b]{\linewidth}\raggedright
\textbf{Pros}
\end{minipage} & \begin{minipage}[b]{\linewidth}\raggedright
\textbf{Cons}
\end{minipage} \\
\midrule()
\endhead
{[}63{]}, {[}64{]}, {[}65{]} & Virtual reality, memory server, basic
processor. & \begin{minipage}[t]{\linewidth}\raggedright
\begin{itemize}
\item
  Patient care
\item
  Entertainment
\item
  Surgeon and nurse training
\item
  dental implant training
\item
  Nurse transition
\end{itemize}
\end{minipage} & Suitable for medical training and patient
entertainment, smart virtual reality & Not useful for disease detection,
skilled persons are required to handle \\
{[}71{]}, {[}67{]} & Microsoft Kinect, and other bio-signal capture
devices, Virtual reality, high-end processors, and memory servers. &
\begin{minipage}[t]{\linewidth}\raggedright
\begin{itemize}
\item
  Neuro-psychological rehabilitation
\item
  upper limb rehabilitation
\item
  chronic pain rehabilitation diagnosis
\item
  management of Parkinson's
\end{itemize}
\end{minipage} & 3D versions are available & Not user-friendly, EEG
signal analysis complexity, Expensive, Expert/skilled person required \\
{[}68{]}, {[}70{]} & Virtual reality, NLP algorithms, IoT, Hypertext
Application Language, MySQL, taxonomy, ML APIs. &
\begin{minipage}[t]{\linewidth}\raggedright
\begin{itemize}
\item
  Fat and carbohydrate Oxidation
\item
  heart rate improvement
\item
  energy expenditure
\end{itemize}
\end{minipage} & Smart GUI, scalability, auto adaptability,
virtualization & the use of Node-RED and other third-party APIs,
Expensive, Expert/skilled person required \\
{[}69{]} & Blending reality and AR/VR, NVIDIA DLSS algorithm, Mobile
devices & \begin{minipage}[t]{\linewidth}\raggedright
\begin{itemize}
\item
  Health monitoring
\item
  clinical detection, treatment, rehabilitation
\item
  recovery tracking, professional
\end{itemize}
\end{minipage} & Intelligent graphics design, 3D application &
Expensive, needs skilled users to handle \\
{[}66{]} & Bayesian game-theoretic model, Multi-objective Evolutionary
Algorithm using Fuzzy Genetics &
\begin{minipage}[t]{\linewidth}\raggedright
\begin{itemize}
\item
  Clinical Decision Support System
\end{itemize}
\end{minipage} & Efficient resource utilization, and computationally
cost-effective solution & Low Accuracy \\
\textbf{Proposed}

\textbf{MOD-CNN} & 1D and 2D CNN deep learning architectures for two
layers of training \& testing, Javascript, HTML \& CSS, memory server,
OS, basic hardware, Mobile devices (flexible) &
\begin{minipage}[t]{\linewidth}\raggedright
\begin{itemize}
\item
  Dementia detection of gaming application users
\end{itemize}
\end{minipage} & First-of-its-kind, easy-to-install, dementia detection
testing accuracy is high, portable, reliable, secure, quickly
responsive, user-friendly, age-independent, and cost-effective & 2D
Gaming application needs to improve the layout \\
\bottomrule()
\end{longtable}

Table 15 provides a comparison of our proposed AI-based gaming
application with some existing state-of-the-art healthcare gaming
application works concerning used technology, application areas,
advantages, and disadvantages.

\textbf{5. CONCLUSIONS}

In this work, we have introduced an innovative cognitive assessment game
that seamlessly integrates gameplay elements with a sophisticated early
diagnosis methodology. The game comprises two levels - ``Basic Match''
and ``Ultimate Test'' and evaluates critical cognitive abilities that
are essential for early dementia identification. The transition from
gameplay to cognitive assessment is smooth, enhancing the overall
utility of the proposed approach. The research incorporates two models
tailored for the binary classification of dementia status using a health
metrics dataset as well as a facial image dataset. The introduced
Modified 1D Convolutional Neural Network (MOD-1D-CNN) is used to analyze
a dataset comprising health metrics for accurately identifying dementia
traits and the other model i.e. Modified 2D Convolutional Neural Network
(MOD-2D-CNN) introduces a facial-based image classification for the
demented as well non-demented individual. Through rigorous evaluation
and comparison-based analysis with the other state-of-the-art models,
the MOD-1D-CNN and MOD-2D-CNN models show their superiority in accuracy,
efficiency, and generalization capabilities, solidifying their position
as noble and valuable contributions to the field of dementia detection.
Both the models with the efficient parameter design ensure practicality
and feasibility in resource-constrained settings, including regions with
limited computational resources. Moreover, the MOD-1D-CNN and MOD-2D-CNN
model's interpretability fosters transparency and enhances trust
enabling them to make more informed decisions based on model
predictions. The work contributes to bridging the research gap in early
dementia detection by presenting a holistic cognitive assessment system
that amalgamates gaming, data-driven analytics, and advanced deep
learning techniques. The findings underscore the potential of this
approach for early identification of cognitive deviations, thereby
advancing the field of dementia research and cognitive health
assessment.

\textbf{ACKNOWLEDGEMENT}

We have obtained the health parameter datasets from Apollo Clinic,
Kolkata. We appreciate Mr. Souvik Das, Floor Manager, Apollo Clinic,
Kolkata, for dedicating his time, effort, and continuous assistance in
gathering the dataset.

\textbf{Ethical Approval}

Not Applicable.

\textbf{Competing Interest}

The authors declare no conflict of interest.

\textbf{Consent to Participate}

All the authors have equally participated in this work.

\textbf{Funding}

There is no funding involved for this work.

\textbf{Data Availability Statement}

Data are available upon reasonable request to the corresponding authors.

\textbf{REFERENCES}

{[}1{]} Valcour, V. G., Masaki, K. H., Curb, J. D., \& Blanchette, P. L.
(2000). The detection of dementia in the primary care setting. Archives
of internal medicine, 160(19), 2964-2968.

{[}2{]} Giaquinto, S., \& Parnetti, L. (2006). Early detection of
dementia in clinical practice. Mechanisms of ageing and development,
127(2), 123-128.

{[}3{]} Palmer, K., Bäckman, L., Winblad, B., \& Fratiglioni, L. (2003).
Detection of Alzheimer's disease and dementia in the preclinical phase:
population based cohort study. Bmj, 326(7383), 245.

{[}4{]} Graafland, M., \& Schijven, M. (2018). How serious games will
improve healthcare. \emph{Digital health: scaling healthcare to the
world}, 139-157.

{[}5{]} Raja, B. S., \& Asghar, S. (2020). Disease classification in
health care systems with game theory approach. \emph{IEEE Access},
\emph{8}, 83298-83311.

{[}6{]} Abd-Alrazaq, A., Abuelezz, I., Hassan, A., AlSammarraie, A.,
Alhuwail, D., Irshaidat, S., ... \& Househ, M. (2022). Artificial
Intelligence--Driven Serious Games in Health Care: Scoping Review.
\emph{JMIR serious games}, \emph{10}(4), e39840.

{[}7{]} Kherchouche, A., Ben-Ahmed, O., Guillevin, C., Tremblais, B.,
Julian, A., Fernandez-Maloigne, C., \& Guillevin, R. (2022).
Attention-guided neural network for early dementia detection using MRS
data. \emph{Computerized Medical Imaging and Graphics}, \emph{99},
102074.

{[}8{]} Alex, J. S. R., Bhyri, R., Usha, G. P., \& Arvind, S. V. (2023).
Noninvasive Detection of Alzheimer's Disease from Conversational Speech
Using 1D-CNN. In \emph{Futuristic Communication and Network
Technologies: Select Proceedings of VICFCNT 2021, Volume 1} (pp.
583-592). Singapore: Springer Nature Singapore.

{[}9{]} Qayyum, A., Khan, M. A., Mazher, M., \& Suresh, M. (2018,
November). Classification of eeg learning and resting states using
1d-convolutional neural network for cognitive load assesment. In
\emph{2018 IEEE Student Conference on Research and Development (SCOReD)}
(pp. 1-5). IEEE.

{[}10{]} Paliokas, I., Tzallas, A., Katertsidis, N., Votis, K., \&
Tzovaras, D. (2017, October). Gamification in social networking: a
platform for people living with dementia and their caregivers. In
\emph{2017 IEEE 17th International Conference on Bioinformatics and
Bioengineering (BIBE)} (pp. 574-579). IEEE.

{[}11{]} Zhang, B., \& Chignell, M. (2020, August). A framework for
using cognitive assessment games for people living with dementia. In
\emph{2020 IEEE 8th International Conference on Serious Games and
Applications for Health (SeGAH)} (pp. 1-8). IEEE.

{[}12{]} Chi, H., Agama, E., \& Prodanoff, Z. G. (2017, April).
Developing serious games to promote cognitive abilities for the elderly.
In \emph{2017 IEEE 5th International Conference on Serious Games and
Applications for Health (SeGAH)} (pp. 1-8). IEEE.

{[}13{]} Mora, A., González, C., Arnedo-Moreno, J., \& Álvarez, A.
(2016, September). Gamification of cognitive training: a
crowdsourcing-inspired approach for older adults. In \emph{Proceedings
of the XVII international conference on human computer interaction} (pp.
1-8).

{[}14{]} Paletta, L., Pszeida, M., Fellner, M., Russegger, S., Dini, A.,
Draxler, S., ... \& Steiner, J. (2021). Towards Decision Support with
Assessment of Neuropsychological Profiles in Alzheimer's Dementia Using
Playful Tablet-Based Multimodal Activation. In \emph{Advances in
Neuroergonomics and Cognitive Engineering: Proceedings of the AHFE 2021
Virtual Conferences on Neuroergonomics and Cognitive Engineering,
Industrial Cognitive Ergonomics and Engineering Psychology, and
Cognitive Computing and Internet of Things, July 25-29, 2021, USA} (pp.
469-477). Springer International Publishing.

{[}15{]} So, A., Hooshyar, D., Park, K. W., \& Lim, H. S. (2017). Early
diagnosis of dementia from clinical data by machine learning techniques.
\emph{Applied Sciences}, \emph{7}(7), 651.

{[}16{]} Shankle, W. R., Mani, S., Pazzani, M. J., \& Smyth, P. (1997).
Detecting very early stages of dementia from normal aging with machine
learning methods. In \emph{Artificial Intelligence in Medicine: 6th
Conference on Artificial Intelligence in Medicine Europe, AIME'97
Grenoble, France, March 23--26, 1997 Proceedings 6} (pp. 71-85).
Springer Berlin Heidelberg.

{[}17{]} Bidani, A., Gouider, M. S., \& Travieso-González, C. M. (2019).
Dementia detection and classification from MRI images using deep neural
networks and transfer learning. In \emph{Advances in Computational
Intelligence: 15th International Work-Conference on Artificial Neural
Networks, IWANN 2019, Gran Canaria, Spain, June 12-14, 2019,
Proceedings, Part I 15} (pp. 925-933). Springer International
Publishing.

{[}18{]} Murugan, S., Venkatesan, C., Sumithra, M. G., Gao, X. Z.,
Elakkiya, B., Akila, M., \& Manoharan, S. (2021). DEMNET: a deep
learning model for early diagnosis of Alzheimer diseases and dementia
from MR images. \emph{Ieee Access}, \emph{9}, 90319-90329.

{[}19{]} Kevin S. McGrew, CHC theory and the human cognitive abilities
project: Standing on the shoulders of the giants of psychometric
intelligence research, Intelligence, Volume 37, Issue 1, 2009, Pages
1-10, ISSN 0160-2896.

{[}20{]} Funahashi S, Andreau JM. Prefrontal cortex and neural
mechanisms of executive function. J Physiol Paris. 2013
Dec;107(6):471-82. doi: 10.1016/j.jphysparis.2013.05.001. Epub 2013 May
15. PMID: 23684970.

{[}21{]} Heatherton TF, Wagner DD. Cognitive neuroscience of
self-regulation failure. Trends Cogn Sci. 2011 Mar;15(3):132-9. doi:
10.1016/j.tics.2010.12.005. Epub 2011 Jan 26. PMID: 21273114; PMCID:
PMC3062191.

{[}22{]} Gonzalez Kelso I, Tadi P. Cognitive Assessment. 2022 Nov 7. In:
StatPearls {[}Internet{]}. Treasure Island (FL): StatPearls Publishing;
2024 Jan--. PMID: 32310509.

{[}23{]} Basu, Saikat \& Saha, Sudipta \& Das, Sourav \& Guha, Rajlaksmi
\& Mukherjee, Jayanta \& Mahadevappa, Manjunatha. (2022). ``Montaj'': A
Gaming System for Assessing Cognitive Skills in a Mobile Computing
Platform. 10.1007/978-981-19-1408-9\_12.

{[}24{]} Basu, Saikat \& Saha, Sudipta \& Das, Sourav \& Guha, Rajlaksmi
\& Mukherjee, Jayanta \& Mahadevappa, Manjunatha. (2022). Assessment of
attention and working memory among young adults using computer games.
Journal of Ambient Intelligence and Humanized Computing. 14.
10.1007/s12652-022-04494-5.

{[}25{]} O'Shea, K., \& Nash, R. (2015). An introduction to
convolutional neural networks. \emph{arXiv preprint arXiv:1511.08458}.

{[}26{]} Aghdam, H. H., \& Heravi, E. J. (2017). Guide to convolutional
neural networks. \emph{New York, NY: Springer}, \emph{10}(978-973), 51.

{[}27{]} Dakdareh, S. G., \& Abbasian, K. (2024). Diagnosis of
Alzheimer's Disease and Mild Cognitive Impairment Using Convolutional
Neural Networks. \emph{Journal of Alzheimer's Disease Reports},
\emph{8}(1), 317.

{[}28{]} Kiranyaz, S., Ince, T., Abdeljaber, O., Avci, O., \& Gabbouj,
M. (2019, May). 1-D convolutional neural networks for signal processing
applications. In \emph{ICASSP 2019-2019 IEEE International Conference on
Acoustics, Speech and Signal Processing (ICASSP)} (pp. 8360-8364). IEEE.

{[}29{]} Kiranyaz, S., Avci, O., Abdeljaber, O., Ince, T., Gabbouj, M.,
\& Inman, D. J. (2021). 1D convolutional neural networks and
applications: A survey. \emph{Mechanical systems and signal processing},
\emph{151}, 107398.

{[}30{]} Li, D., Zhang, J., Zhang, Q., \& Wei, X. (2017, October).
Classification of ECG signals based on 1D convolution neural network. In
\emph{2017 IEEE 19th International Conference on e-Health Networking,
Applications and Services (Healthcom)} (pp. 1-6). IEEE.

{[}31{]} Mattioli, F., Porcaro, C., \& Baldassarre, G. (2022). A 1D CNN
for high accuracy classification and transfer learning in motor imagery
EEG-based brain-computer interface. \emph{Journal of Neural
Engineering}, \emph{18}(6), 066053.

{[}32{]} Li, F., Liu, M., Zhao, Y., Kong, L., Dong, L., Liu, X., \& Hui,
M. (2019). Feature extraction and classification of heart sound using 1D
convolutional neural networks. \emph{EURASIP Journal on Advances in
Signal Processing}, \emph{2019}(1), 1-11.

{[}33{]} Allamy, S., \& Koerich, A. L. (2021, December). 1D CNN
architectures for music genre classification. In \emph{2021 IEEE
Symposium Series on Computational Intelligence (SSCI)} (pp. 01-07).
IEEE.

{[}34{]} Hsieh, C. H., Li, Y. S., Hwang, B. J., \& Hsiao, C. H. (2020).
Detection of atrial fibrillation using 1D convolutional neural network.
\emph{Sensors}, \emph{20}(7), 2136.

{[}35{]} Hu, H., Shah, S. A. A., Bennamoun, M., \& Molton, M. (2017,
November). 2D and 3D face recognition using convolutional neural
network. In \emph{TENCON 2017-2017 IEEE Region 10 Conference} (pp.
133-132). IEEE.

{[}36{]} Taha, B., \& Hatzinakos, D. (2019, May). Emotion recognition
from 2D facial expressions. In \emph{2019 IEEE Canadian Conference of
Electrical and Computer Engineering (CCECE)} (pp. 1-4). IEEE.

{[}37{]} Li, H., Sun, J., Xu, Z., \& Chen, L. (2017). Multimodal 2D+ 3D
facial expression recognition with deep fusion convolutional neural
network. \emph{IEEE Transactions on Multimedia}, \emph{19}(12),
2816-2831.

{[}38{]} TanujaPatgar, T. (2021). Convolution neural network based
emotion classification cognitive modelforfacial expression.
\emph{Turkish Journal of Computer and Mathematics Education (TURCOMAT)},
\emph{12}(13), 6718-6739.

{[}39{]} Almabdy, S., \& Elrefaei, L. (2019). Deep convolutional neural
network-based approaches for face recognition. \emph{Applied Sciences},
\emph{9}(20), 4397.

{[}40{]} Wu, C., \& Guo, F. (2021). TSNN: Three‐stream combining 2D and
3D convolutional neural network for micro‐expression recognition.
\emph{IEEJ Transactions on Electrical and Electronic Engineering},
\emph{16}(1), 98-107.

{[}41{]} Cui, J., Zhang, H., Han, H., Shan, S., \& Chen, X. (2018,
February). Improving 2D face recognition via discriminative face depth
estimation. In \emph{2018 International Conference on Biometrics (ICB)}
(pp. 140-147). IEEE.

{[}42{]} Li, J., Mi, Y., Li, G., \& Ju, Z. (2019). CNN-based facial
expression recognition from annotated rgb-d images for human--robot
interaction. \emph{International Journal of Humanoid Robotics},
\emph{16}(04), 1941002.

{[}43{]} Kim, H., Jung, W. K., Park, Y. C., Lee, J. W., \& Ahn, S. H.
(2022). Broken stitch detection method for sewing operation using CNN
feature map and image-processing techniques. \emph{Expert Systems with
Applications}, \emph{188}, 116014.

{[}44{]} Menikdiwela, M., Nguyen, C., Li, H., \& Shaw, M. (2017,
December). CNN-based small object detection and visualization with
feature activation mapping. In \emph{2017 international conference on
image and vision computing New Zealand (IVCNZ)} (pp. 1-5). IEEE.

{[}45{]} Singh, S., \& Anand, R. S. (2019). Multimodal medical image
fusion using hybrid layer decomposition with CNN-based feature mapping
and structural clustering. \emph{IEEE Transactions on Instrumentation
and Measurement}, \emph{69}(6), 3855-3865.

{[}46{]} Guyon, I., \& Elisseeff, A. (2006). An introduction to feature
extraction. In Feature extraction: foundations and applications (pp.
1-25). Berlin, Heidelberg: Springer Berlin Heidelberg.

{[}47{]} Wu, B. F., \& Lin, C. H. (2018). Adaptive feature mapping for
customizing deep learning based facial expression recognition model.
\emph{IEEE access}, \emph{6}, 12451-12461.

{[}48{]} Ge, Z., Cao, G., Li, X., \& Fu, P. (2020). Hyperspectral image
classification method based on 2D--3D CNN and multibranch feature
fusion. \emph{IEEE Journal of Selected Topics in Applied Earth
Observations and Remote Sensing}, \emph{13}, 5776-5788.

{[}49{]} Steurer, M., Hill, R. J., \& Pfeifer, N. (2021). Metrics for
evaluating the performance of machine learning based automated valuation
models. \emph{Journal of Property Research}, \emph{38}(2), 99-129.

{[}50{]} Albawi, S., Mohammed, T. A., \& Al-Zawi, S. (2017, August).
Understanding of a convolutional neural network. In \emph{2017
international conference on engineering and technology (ICET)} (pp.
1-6). Ieee.

{[}51{]} Zhou, J., Gandomi, A. H., Chen, F., \& Holzinger, A. (2021).
Evaluating the quality of machine learning explanations: A survey on
methods and metrics. \emph{Electronics}, \emph{10}(5), 593.

{[}52{]} Cui, Y., Zhou, F., Wang, J., Liu, X., Lin, Y., \& Belongie, S.
(2017). Kernel pooling for convolutional neural networks. In
\emph{Proceedings of the IEEE conference on computer vision and pattern
recognition} (pp. 2921-2930).

{[}53{]} Vujović, Ž. (2021). Classification model evaluation metrics.
\emph{International Journal of Advanced Computer Science and
Applications}, \emph{12}(6), 599-606.

{[}54{]} Hao, W., Yizhou, W., Yaqin, L., \& Zhili, S. (2020, December).
The role of activation function in CNN. In \emph{2020 2nd International
Conference on Information Technology and Computer Application (ITCA)}
(pp. 429-432). IEEE.

{[}55{]} Hossin, M., \& Sulaiman, M. N. (2015). A review on evaluation
metrics for data classification evaluations. \emph{International journal
of data mining \& knowledge management process}, \emph{5}(2), 1.

{[}56{]} Park, S., \& Kwak, N. (2017). Analysis on the dropout effect in
convolutional neural networks. In \emph{Computer Vision--ACCV 2016: 13th
Asian Conference on Computer Vision, Taipei, Taiwan, November 20-24,
2016, Revised Selected Papers, Part II 13} (pp. 189-204). Springer
International Publishing.

{[}57{]} Carvalho, D. V., Pereira, E. M., \& Cardoso, J. S. (2019).
Machine learning interpretability: A survey on methods and metrics.
\emph{Electronics}, \emph{8}(8), 832.

{[}58{]} Jin, J., Dundar, A., \& Culurciello, E. (2014). Flattened
convolutional neural networks for feedforward acceleration. \emph{arXiv
preprint arXiv:1412.5474}.

{[}59{]} Bera, S., \& Shrivastava, V. K. (2020). Analysis of various
optimizers on deep convolutional neural network model in the application
of hyperspectral remote sensing image classification.
\emph{International Journal of Remote Sensing}, \emph{41}(7), 2664-2683.

{[}60{]} Jha, S., Kumar, R., Abdel-Basset, M., Priyadarshini, I.,
Sharma, R., \& Long, H. V. (2019). Deep learning approach for software
maintainability metrics prediction. \emph{Ieee Access}, \emph{7},
61840-61855.

{[}61{]} Marcus, D. S., Fotenos, A. F., Csernansky, J. G., Morris, J.
C., \& Buckner, R. L. (2010). Open access series of imaging studies:
longitudinal MRI data in nondemented and demented older adults.
\emph{Journal of cognitive neuroscience}, \emph{22}(12), 2677-2684.

{[}62{]} Marom, N. D., Rokach, L., \& Shmilovici, A. (2010, November).
Using the confusion matrix for improving ensemble classifiers. In
\emph{2010 IEEE 26-th Convention of Electrical and Electronics Engineers
in Israel} (pp. 000555-000559). IEEE.

{[}63{]} Wattanasoontorn, V., Hernández, R. J. G., \& Sbert, M. (2014).
Serious games for e-health care. \emph{Simulations, Serious Games and
Their Applications}, 127-146.

{[}64{]} Kato, P. M. (2010). Video games in health care: Closing the
gap. \emph{Review of general psychology}, \emph{14}(2), 113-121.

{[}65{]} Abd-Alrazaq, A., Abuelezz, I., Hassan, A., AlSammarraie, A.,
Alhuwail, D., Irshaidat, S., ... \& Househ, M. (2022). Artificial
Intelligence--Driven Serious Games in Health Care: Scoping Review.
\emph{JMIR serious games}, \emph{10}(4), e39840.

{[}66{]} Raja, B. S., \& Asghar, S. (2020). Disease classification in
health care systems with game theory approach. \emph{IEEE Access},
\emph{8}, 83298-83311.

{[}67{]} Graafland, M., \& Schijven, M. (2018). How serious games will
improve healthcare. \emph{Digital health: scaling healthcare to the
world}, 139-157.

{[}68{]} Ahmad, S., Mehmood, F., Khan, F., \& Whangbo, T. K. (2022).
Architecting intelligent smart serious games for healthcare
applications: a technical perspective. \emph{Sensors}, \emph{22}(3),
810.

{[}69{]} Savareh, B. A., \& Bashiri, A. (2021). Artificial intelligence
and Mobile Gaming. \emph{Journal of Biomedical Physics \& Engineering},
\emph{11}.

{[}70{]} Sitterding, M. C., Raab, D. L., Saupe, J. L., \& Israel, K. J.
(2019). Using artificial intelligence and gaming to improve new nurse
transition. \emph{Nurse Leader}, \emph{17}(2), 125-130.

{[}71{]} Damaševičius, R., Maskeliūnas, R., \& Blažauskas, T. (2023).
Serious games and gamification in healthcare: a meta-review.
\emph{Information}, \emph{14}(2), 105.

{[}72{]} https://www.pexels.com/, Last Access Date: 15.03.2024.

{[}73{]} \url{https://create.vista.com/}, Last Access Date: 15.03.2024.

{[}74{]} \url{https://images.google.com/}, Last Access Date: 15.03.2024.

{[}75{]} \url{https://www.shutterstock.com/}, Last Access Date:
15.03.2024.

\end{document}